\documentclass[conference]{IEEEtran}
\IEEEoverridecommandlockouts
\usepackage{cite}
\usepackage{amsmath,amssymb,amsfonts}
\usepackage{algorithmic}
\usepackage{graphicx}
\usepackage{textcomp}
\usepackage{xcolor}
\usepackage[colorlinks,linkcolor=blue]{hyperref}
\usepackage{amsmath,amssymb,amsfonts}

\usepackage{tikz}
\usepackage{comment}
\usepackage{amsmath,amssymb} 
\usepackage{color}
\usepackage{xcolor}
\usepackage{graphicx}
\usepackage{subfigure}
\usepackage{algorithm}
\usepackage{algorithmic}
\usepackage{booktabs}
\usepackage{multirow}
\usepackage[accsupp]{axessibility}  
\usepackage{amsthm}
\newtheorem{assumption}{Assumption}
\newtheorem{theorem}{Theorem}

\newtheorem{lemma}{Lemma}

\usepackage{amsmath,amsfonts,bm}

\def\eqref#1{eqn.~(\ref{#1})}

\def\1{\bm{1}}

\DeclareMathAlphabet{\mathsfit}{\encodingdefault}{\sfdefault}{m}{sl}
\SetMathAlphabet{\mathsfit}{bold}{\encodingdefault}{\sfdefault}{bx}{n}

\def\g{{\bf g}}

\def\tt{{\bf t}}

\def\u{{\bf u}}

\def\v{{\bf v}}

\def\w{{\bf w}}

\def\0{{\bf 0}}
\def\1{{\bf 1}}

\def\EB{{\mathbb E}}

\newcommand{\aw}{\overline{\w}}

\newcommand{\myave}{\sum_{k=1}^N p_k}

\newcommand{\av}{\overline{\v}}

\newcommand{\red}[1]{\textcolor{black}{#1}}
\newcommand{\rednew}[1]{\textcolor{black}{#1}}

\def\BibTeX{{\rm B\kern-.05em{\sc i\kern-.025em b}\kern-.08em
    T\kern-.1667em\lower.7ex\hbox{E}\kern-.125emX}}
\begin{document}

\title{FedSkip: Combatting Statistical Heterogeneity with Federated Skip Aggregation}

\author{\IEEEauthorblockN{Ziqing Fan}
\thanks{* Corresponding Authors.}
\IEEEauthorblockA{\textit{CMIC, Shanghai Jiao Tong University} \\
\textit{Shanghai AI Laboratory}\\
Shanghai, China \\
zqfan\_knight@sjtu.edu.cn
}
\and
\IEEEauthorblockN{Yanfeng Wang*}
\IEEEauthorblockA{\textit{CMIC, Shanghai Jiao Tong University} \\
\textit{Shanghai AI Laboratory}\\
Shanghai, China \\
wangyanfeng@sjtu.edu.cn}
\and
\IEEEauthorblockN{Jiangchao Yao*}
\IEEEauthorblockA{\textit{CMIC, Shanghai Jiao Tong University} \\
\textit{Shanghai AI Laboratory}\\
Shanghai, China \\
sunarker@sjtu.edu.cn}
\and
\IEEEauthorblockN{Lingjuan Lyu}
\IEEEauthorblockA{\textit{Sony AI} \\
Japan \\
lingjuanlvsmile@gmail.com}
\and
\IEEEauthorblockN{Ya Zhang}
\IEEEauthorblockA{\textit{CMIC, Shanghai Jiao Tong University} \\
\textit{Shanghai AI Laboratory}\\
Shanghai, China \\
ya\_zhang@sjtu.edu.cn}
\and
\IEEEauthorblockN{Qi Tian}
\IEEEauthorblockA{\textit{Huawei Cloud \& AI} \\
China \\
tian.qi1@huawei.com}
}


\maketitle
\begin{abstract}
The statistical heterogeneity of the non-\emph{independent and identically distributed} (non-IID) data in local clients significantly limits the performance of federated learning. \red{Previous attempts like FedProx, SCAFFOLD, MOON, FedNova and FedDyn resort to an optimization perspective, which requires an auxiliary term or re-weights local updates to calibrate the learning bias or the objective inconsistency.} However, \rednew{in addition to previous explorations for improvement in federated averaging, our analysis shows that another critical bottleneck is the poorer optima of client models in more heterogeneous conditions.} We thus introduce a data-driven approach called FedSkip to improve the client optima by periodically skipping federated averaging and scattering local models to the cross devices. We provide theoretical analysis 
of the possible benefit from FedSkip and conduct extensive experiments on a range of datasets to demonstrate that FedSkip achieves much higher accuracy, better aggregation efficiency \red{and competing communication efficiency}s. Source code
is available at:~\href{https://github.com/MediaBrain-SJTU/FedSkip}{https://github.com/MediaBrain-SJTU/FedSkip}.
\end{abstract}


\section{Introduction}
Federated learning~\cite{fedavg} has drawn the considerable attention in recent years due to the increasing concerns about data privacy~\cite{privacy1,privacy2,privacy3,privacy4}. A range of real-world applications such as medical diagnosis~\cite{medicalimage1,medicalimage2}, autonomous driving~\cite{driving1,driving2} and visual recognition~\cite{liu2020fedvision,hsu2020federated} have applied federated learning 
to avoid the potential privacy violation. Nevertheless, there are several remaining challenges on heterogeneity~\cite{noniid2} and efficiency~\cite{effi2} that need to be solved for practical pursuits.

As a distributed learning paradigm, federated learning trains the model in the local without directly communicating the raw data~\cite{konevcny2016federated}. Specifically, the representative optimization method in federated learning, FedAvg~\cite{fedavg}, reduces the parameter transmission cost significantly by increasing the steps of the local training. Despite communication effectiveness, the statistical heterogeneity due to the non-IID nature among clients greatly limits the performance of federated learning~\cite{noniid1,noniid2}. To tackle this challenge, many attempts have been made to improve the federated optimization~\cite{fedprox,scaffold,moon,fednova,fedyn}. For example, FedNova~\cite{fednova} introduces a normalizing weight to eliminate the objective inconsistency, and FedProx~\cite{fedprox}, SCAFFOLD~\cite{scaffold}, MOON~\cite{moon} \red{and FedDyn~\cite{fedyn}} explore to reduce the variance or calibrate the optimization in FedAvg by adding the proximal regularization. All these methods mainly focus on minimizing the discrepancy between the global model and the client models during the local training or the global aggregation.

\begin{figure*}[t]
    \centering
    \subfigure[FedAvg under different heterogeneity.
    ]{
    \centering
    \label{fig:problem}
    \includegraphics[width=0.42\textwidth]{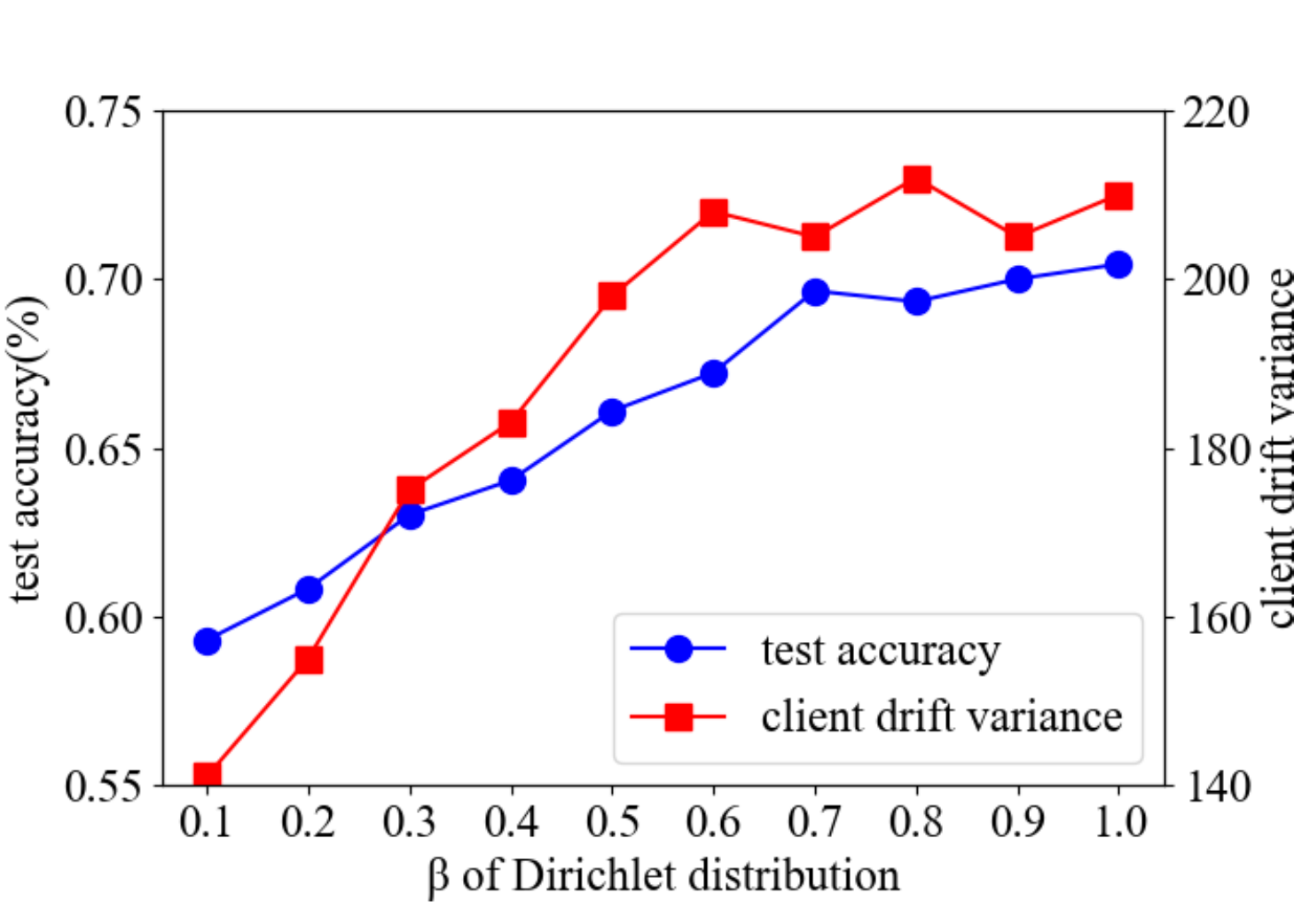}}
    \hspace{2mm}
    \subfigure[Illustration about the phenomenon.]{
    \centering
    \label{fig:moti}
    \includegraphics[width=0.41\textwidth]{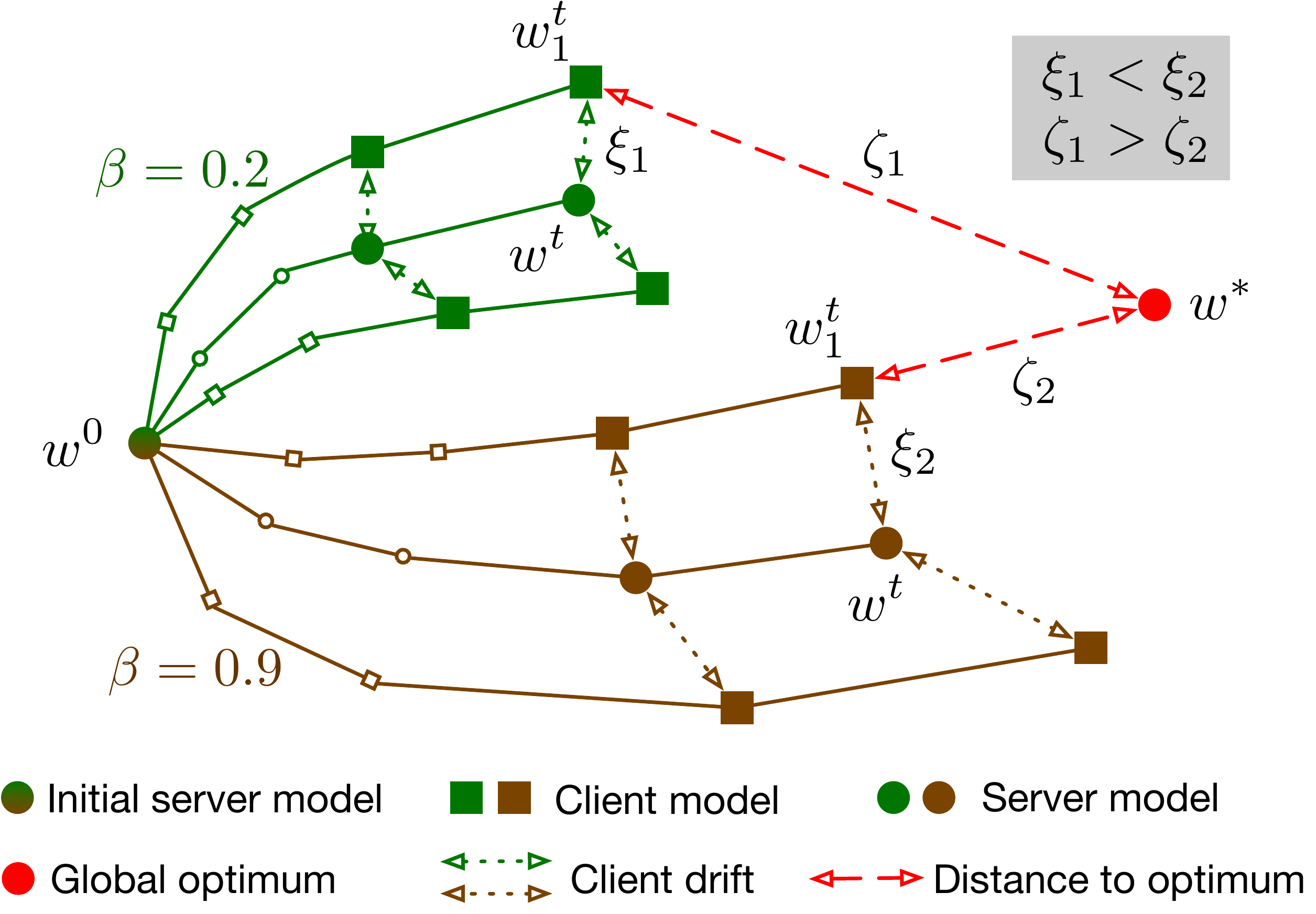}}
    \caption{The analysis about FedAvg under different heterogeneity on CIFAR-10. In the left panel, we plot the performance of FedAvg and the corresponding client drift of FedAvg under different heterogeneous partitions of CIFAR-10 (smaller $\beta$, more heterogeneous). In the right panel, we illustrate the potential explanation behind the left panel through two types of heterogeneity and their averaging between two clients. Namely, comparing with the less heterogeneous condition ($\beta=0.9$), in a more heterogeneous condition ($\beta=0.2$), we have a smaller drift variance ($\xi_1 < \xi_2$) but a worse client optimum ($\zeta_1>\zeta_2$).}\label{fig:motivation}
\end{figure*}

\red{However, in Fig.~\ref{fig:problem}, we observe one interesting trend that the more heterogeneous (indicated by the smaller Dirichlet parameter $\beta$ \cite{dili1,noniid2,dili2,dili3}) the conditions are, the smaller the client drift variance (\textit{i.e.,} smaller discrepancy between the global model and the client models) they have.}  We conjecture that the potential reason is like Fig.~\ref{fig:moti}. Namely, after the periodical federated aggregation on the server side, the local models in more heterogeneous scenarios are more easily stuck near the server model with the smaller variance but unfortunately far away from the desired global optimum. Although minimizing the discrepancy in the general case like FedDyn, FedProx, SCAFFOLD and MOON can be helpful, we shall notice that another critical bottleneck is actually how to improve the client optima to approach the global optimum. 

Based on the above analysis, we propose a novel data-driven approach, termed as FedSkip, which aims to improve the client optima instead of the intuition in the aforementioned methods. Specifically, in FedSkip, the central server periodically skips the federated aggregation and directly scatters the received client models to the cross clients. This routing through the scattering operation allows the local model to access the data in multiple clients for a better optimum pursuit. Note that, here we randomly shuffle all client models for scattering to avoid the id information leakage. Thus, the local clients cannot reverse the oriented information, which guarantees the scattering operation is at least as safe as FedAvg. Then, in the rounds that the federated averaging is executed, we can aggregate a better server model with the improved client models. We theoretically analyze the convergence characteristics of such a federated skip aggregation operation in federated learning and empirically compare it with FedProx, SCAFFOLD, MOON, FedNova and FedDyn on a range of benchmark datasets. Extensive experimental results demonstrate that the proposed FedSkip has much higher accuracy and aggregation efficiency as well as a competing communication efficiency.

In the following, we will review some related works in Section~\ref{sec:related}. In Section~\ref{sec:delay}, the basic notations and the motivation are presented, followed by the proposed method FedSkip. The corresponding theoretical analysis is shown in Section~\ref{sec:theory}. Then, we introduce the experimental settings and the comparison as well as the ablation study in Section~\ref{sec:experiment}. Finally, we conclude the paper in Section~\ref{sec:conclusion}.

\section{Related Work}\label{sec:related}

\subsection{Federated Learning}
In conventional centralized training paradigms, clients can upload their data to the central server and then a model is learned~\cite{jordan2015machine}. With the recently increasing concern about the data privacy, it becomes a trend to keep the sensitive data in the local~\cite{konevcny2016federated,yao2021device,yao2022edge,yao2022device}. Under this background, federated learning 
emerged and has gained significant attention in many tasks such as medical diagnosis~\cite{medicalimage1,medicalimage2}, auto-driving~\cite{driving1,driving2} and visual recognition~\cite{liu2020fedvision,hsu2020federated}.
\subsection{Statistical Heterogeneity}
Statistical heterogeneity is the characteristic of the non-IID data, which induces the model bias 
in local clients. In terms of traditional centralized training paradigms, this can be automatically avoided by one model via stochastic gradient descent~\cite{verbraeken2020survey}. However, in the context of federated learning settings, statistical heterogeneity becomes a severe problem when aggregating the diverse client models. Specifically, previous studies show that if statistical heterogeneity exists, gradients from different client models are very different, which degrades the performance of federated approaches~\cite{noniid2,noniid1,scaffold,fedprox}.

To tackle the statistical heterogeneity, many solutions have been proposed. A line of research focuses on adding constraints like normalization or regularization on local updates to improve local training. FedProx~\cite{fedprox} reduces bias by adding a proximal term to limit the length of local updates. MOON~\cite{moon} corrects gradients in local training by utilizing the contrastive learning method to decrease the distance between representations learned by local model and global model, and increase the distance between representations learned by local model and previous local model. Another line of research focuses more on eliminating inconsistency and directly correcting updates to improve local models or finding a better way to aggregate. SCAFFOLD~\cite{scaffold} designs a variance reduction approach to achieve a stable and fast local update.  FedNova~\cite{fednova} eliminates the objective inconsistency by normalizing total gradients in one communication round. \red{FedDyn~\cite{fedyn} aligns global and device solutions by proposing a dynamic regularizer for each device at each round.} Different from all these methods, our method FedSkip explores to improve the client optima from the data perspective. By skipping aggregation periodically and scattering local models among clients, FedSkip allows the client model to access more data and \red{maintain the same level of privacy as FedAvg}, 
without needing to apply any constraints or modify aggregation on the server or local training in local clients. 

\section{FedSkip}\label{sec:delay}
In this section, we will first present the basic notations of this work in Section~\ref{sec:priliminary}, and then give our motivation via a simple empirical study in Section~\ref{sec:motivation}, followed by the concrete formulation for FedSkip in Section~\ref{sec:method}.

\subsection{Preliminary}\label{sec:priliminary}
Given a central server and a set of local clients, the classical FedAvg consists of four steps~\cite{fedavg}: 1) In round $t$, the server distributes the global model $\w^{t}$ to local clients that are selected to participate in the training; 2) Each $k$-th local client receives the model ($w^t_k \leftarrow w^t$) and trains the model as follows,
\begin{equation}\label{formula:local}
   \w_{k}^t\leftarrow \w_{k}^t - \eta \nabla F_k(b_{k}^t;\w_{k}^t),
\end{equation}
where $\eta$ is the learning rate, $F_k$ is the loss function of $k$-th local client, $b^{t}_k$ is a mini-batch of training data randomly sampled from the local dataset $\mathcal{D}_k$. After $E$ epochs, we acquire a new local model $\w^{t}_k$;
3) The updated models are then collected to the server as $\{\w^{t}_{1},\w^{t}_{2},\dots,\w^{t}_{K}\}$; 4) the server performs the following aggregation on the collected models to acquire a new global model $\w^{t+1}$,
\begin{equation} \label{eq:fedavg}
    \w^{t+1}\leftarrow \sum_{k=1}^K p_k \w^{t}_{k},
\end{equation}
where $p_k$ is the proportion of sample number of the $k$-th client to the sample number of all the participants, \textit{i.e.,} $p_k=N_k\slash\sum_{k'=1}^K N_{k'}$. When the pre-defined maximal round $T$ reached, we will have the final optimized model $\w^T$. Different from FedAvg, our method will introduce a skip aggregation stage,
\begin{equation} \label{eq:skip}
    I_{skip}=\{~t~|~t\in\mathbb{R}^+~\&\&~1<t< T~\&\&~t~\text{mod}~ \Delta \neq 0\},
\end{equation}
where $\mathbb{R}^+$ is the set of all positive integers, $\Delta$ is a given integer for the skip aggregation, and $t~\text{mod}~ \Delta$ means the mode of $t$ \textit{w.r.t.} $\Delta$. When the round $t\in I_{skip}$, we do perform a novel \emph{federated skip aggregation} instead of federated averaging.

\subsection{Motivation}
As shown in Fig.~\ref{fig:problem}, the smaller $\beta$ 
means a larger heterogeneity 
and a smaller client drift variance~\cite{scaffold}. Note that, the client drift variance is to measure the variance between the global model and the local models. 
That is to say, in the more heterogeneous environment, the local client models are surprisingly homogeneous compared to those in the less heterogeneous environment. The possible explanation is that it is hard to learn the model in more heterogeneous data and thus in each round the updated model 
will not deviate too much from the starting point acquired by initialization or \eqref{eq:fedavg}. Then, the client models fall into the similar local minimums that are far away from the global optimum like Fig.~\ref{fig:moti}. In this case, except for reducing the client drift variance as in~\cite{moon,fedprox}, another critical issue is that how to make client models approach to the global optimum. 

\label{sec:motivation}
Although it is prohibited from transmitting data to the server, we can route local models among clients, which encourages one local model to access more data for training. 
In this way, it is possible to improve the optima of client models. To validate this 
idea, we introduce two training strategies ``Local Training" and ``Cross Training", and conduct a quick toy experiment on CIFAR-10, CIFAR-100 and \red{SHAKESPEARE datasets. CIFAR-10 and CIFAR-100 are divided into 10 local clients under heterogeneous data partitions. SHAKESPEARE is a kind of real world dataset with 660 clients and we randomly select 10 of them.} In Local Training, all client models are respectively trained on their local data for 300 epochs and we compute the average accuracy of client models on the test set. In Cross Training, one model is trained across all clients in a sequential order for 300 epochs, and we compute its accuracy on the test set. \red{This experimental setup can ensure the complete convergence of the two strategies and the details are the same in Section~\ref{sec:experiment}.}

As shown in Table~\ref{tab:prelimi}, we can find that the Cross Training strategy performs much better than the Local Training strategy, which means the local model is closer to the global optimum after routing it among clients. 
Note that, the toy experiments here are to compare the optima in the extremely local training and the sequentially cross-client training, and we have not strictly applied the ``Cross Training" to federate learning, since it does not involve the model averaging in this process. Therefore, a straightforward question is raised here: \emph{Can we build a new method that incorporates the benefit of the Cross Training into federated learning to combat the statistical heterogeneity issue?}

\begin{table}[!t]
    \centering
    \setlength{\tabcolsep}{6pt}
    \small
    \caption{\rm{A toy experiment on CIFAR-10, CIFAR-100 and SHAKESPEARE to compare the 
    performance of ``Local Training" and that of ``Cross Training".}}\label{tab:prelimi}
    \begin{tabular}{ccccccccc}
        \toprule
        Method& CIFAR-10& CIFAR-100& SHAKESPEARE\\
        \midrule
        Local Training &$0.4407$ &$0.2264$ &$0.3088$  \\
        Cross Training &$\textbf{0.5756}$ &$\textbf{0.6005}$ &$\textbf{0.4530}$ \\
        \bottomrule
    \end{tabular}
    
\end{table}

\subsection{The Proposed Approach}
\label{sec:method}
Based on the previous section, we propose a new approach, termed as FedSkip, to combat the statistical heterogeneity of federated learning. In the following, we will present the framework of FedSkip as well as its theoretical analysis. 

\begin{figure*}[!t]
    \centering
    \includegraphics[width=0.98\linewidth]{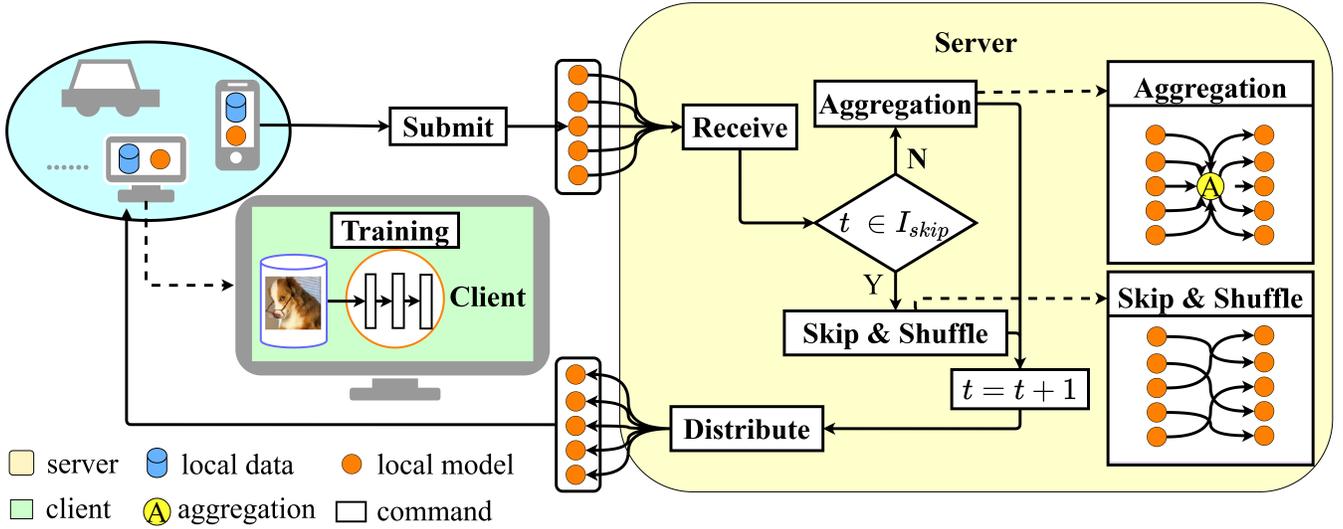}
    \caption{The framework of FedSkip. We illustrate the details of two sides, \textit{i.e.,} the local training in clients~(in green) and federated averaging and federated skip aggregation in the server~(in yellow). The client procedure in FedSkip is same to the original federated learning~\cite{fedavg}, while the server might perform federated averaging (``Aggregation") or federated skip aggregation (``Skip \& Shuffle") based on whether the round $t\in I_{skip}$. }\label{fig:structure}
\end{figure*}

\subsubsection{Framework}
The framework of FedSkip is illustrated in Fig.~\ref{fig:structure}. For the local training step, FedSkip follows the similar procedure as FedAvg where clients receive a model from the server, update it on its private data and then submit the updated model to the server. When the server receives client models, FedSkip, however, has a more flexible step different from FedAvg. Specifically, we additionally introduce a skip stage $I_{skip}$ defined in \eqref{eq:skip} to denote the communication rounds where the server does not perform aggregation. As shown in Fig.~\ref{fig:structure}, for $t \not\in  I_{skip}$, the server will aggregate all client models into a global model and distribute it to local clients as FedAvg. However, if the communication round $t \in I_{skip}$, the server will skip the aggregation and scatter these client models via shuffle to different clients in the next round. 
Consequently, at the beginning of the $t$-th round, the $k$-th client model that participates in the training will receive a different initialization according to the stage
as shown in the following cases,
\begin{equation}
    \label{eq:init}
    \w^{t}_{k}\leftarrow\left\{
    \begin{array}{ll}
\w^{0}, & t=0 \\
{\hat{\w}^{t-1}_{k}} ,  & t~\in~I_{skip}\\
\sum_{k=1}^{K}p_k^t\w_{k}^{t-1}, & \text{otherwise} \\
\end{array} \right.
\end{equation}
where $\hat{\w}^{t-1}_{k}$ is the $k$-th model in $\{\hat{\w}_{1}^{t-1},\hat{\w}_{2}^{t-1},\dots,\hat{\w}_{K}^{t-1}\}$ (the shuffle of $\{\w^{t-1}_{1},$ $\w^{t-1}_{2},\dots,\w^{t-1}_{K}\}$) and $p_k^t$ is the proportion of the number of samples that the $k$-th local model cumulatively uses \textit{w.r.t} the number of samples that all local models cumulatively use in one period of federated skip aggregation, formulated as follows, 
\begin{equation}
p_k^t=\frac{\sum_{n=t-\Delta}^{t}N_{k_n}}{\sum_{k=1}^K{\sum_{n=t-\Delta}^{t}N_{k_n}}}.
\end{equation}
Note that, $N_{k_n}$ is the sample number for the $k$-th model in round $n$. Through \eqref{eq:init}, FedSkip enables the local model to access more than one client's data for training, which improves the optimum of the local model in the heterogeneous environment. 
Moreover, the operation of skipping aggregation and shuffling local models periodically does not break privacy since it happens in the server and each local client do not know where its local model is from and whether it is an averaged model or the skipped model.
This is the key difference of FedSkip from previous methods, a data-driven perspective instead of the optimization perspective~\cite{fedprox,moon,fednova}. In Algorithm {\color{red}\ref{alg:fedavg}}, we give the complete training procedure of FedSkip. At the beginning, local models are initialized by the global model distributed to local clients. After training several epochs on local data as shown in \eqref{formula:local}, local models are submitted to the server. In the server, local models either perform the regular averaging (``Aggregation") or skip aggregation (``Skip $\&$ Shuffle")
depending on the round $t$. 

\begin{algorithm}[!t]
    \caption{FedSkip}
    \label{alg:fedavg}
    \textbf{Input}: a set of $K$ clients 
    that participate into the training in each round, the initializing model weight $\w^{0}$, the learning rate $\eta$, the maximal round $T$, the local training epochs $E$, the pre-defined skip aggregation stage $I_{skip}$.
    \begin{algorithmic}
    \FOR{$t = 0,1,\dots,T-1$}
        \STATE \colorbox{gray!30}{$\rhd$ on the server side}
        \STATE randomly sample $K$ clients to participate into the training.
        \IF{$t=0$}
            \STATE $\forall{k},~\w_k^t \leftarrow \w^0$.
        \ELSIF{$t \in I_{skip}$}
            \STATE shuffle $\{\w_{1}^{t-1},\w_{2}^{t-1},\dots,\w_{K}^{t-1}\}$ as $\{\hat{\w}_{1}^{t-1},\hat{\w}_{2}^{t-1},\dots,\hat{\w}_{K}^{t-1}\}$.
            \STATE $\forall{k},~\w_k^t \leftarrow \hat{\w}_k^{t-1}$.
        \ELSE
            \STATE $p_k^t=\frac{\sum_{n=t-\Delta}^{t}N_{k_n}}{\sum_{k=1}^K{\sum_{n=t-\Delta}^{t}N_{k_n}}}$
            \STATE $\w^{t}\leftarrow \sum_{k=1}^{K}{p^t_k\w^{t-1}_{k}}$.
            \STATE $\forall{k},~\w_k^t \leftarrow \w^t$.
        \ENDIF
        \STATE distribute each $\w_k^t$ to the corresponding $k$-th client.
        \STATE \colorbox{gray!30}{$\rhd$ on the client side}
        \ONCLIENT{$\forall{k \in K}$}
            \FOR{$\tau=(t-1)E,(t-1)E+1,...,tE-1$} 
                \STATE sample a mini-batch $b_k^\tau$ from the local dataset $D_k$. 
                \STATE $\w^{t}_{k}\leftarrow \w^{t}_{k}-\eta\nabla F(b_k^\tau;\w^{t}_{k}) $.
            \ENDFOR
        \STATE submit $\w^{t}_{k}$ to the server.
        \ENDON
    \ENDFOR
    \end{algorithmic}
\end{algorithm}

\section{Theoretical Analysis}~\label{sec:theory}
In this part, we will provide the theoretical analysis of FedSkip. Concretely, we first define several assumptions in Section~\ref{sec:assumption}, which follows similar conditions for FedAvg~\cite{convergence,lemma4}. We then provide the convergence guarantee of FedSkip in Section~\ref{sec:convergence}. The complete proof is given in the Appendix. Finally, we compare the time complexity and the space complexity of FedSkip and FedAvg in Section~\ref{sec:complexity}. 

\subsection{Notation and Assumptions}\label{sec:assumption}
First, we make the assumptions on the loss functions~$F_1, F_2,\cdots, F_N$ of all clients and their gradient functions~$\nabla F_1, \nabla F_2,\cdots, \nabla F_N$. In Assumption~\ref{asm:smooth_and_strong_cvx}, we characterize the smoothness and the convexity of each $F_k$. Assumption~\ref{asm:sgd_var_and_norm} bounds the variance and the norm of stochastic gradients. Here, we denote $\tau$ as the number of the accumulative local updates (from the beginning of federated learning) and $\w_k^\tau$ as the weights of $k$-th local model with $\tau$ rounds of local updates.
\begin{assumption}[L-smooth and $\mu$-strongly convex] \label{asm:smooth_and_strong_cvx}
        $F_1, \cdots, F_N$ are all $L$-smooth and $\mu$-strongly convex:
        for all $\u$ and $\v$, we have $\frac{\mu }{2} \| \u - \v\|_2^2 \leq F_k(\u)  - F_k(\v) - (\u - \v)^T \nabla F_k(\v) \leq  \frac{L}{2} \| \u - \v\|_2^2$.
\end{assumption}

\begin{assumption}[Bounded variance and norm of stochastic gradients] \label{asm:sgd_var_and_norm}
        The variance and the expected squared norm of stochastic gradients are both bounded: for all $k=1,\cdots,N$ and $\tau\bmod{E}\neq0$, we have $\EB \left\| \nabla F_k(\w^\tau_k,\xi^\tau_k) - \nabla F_k(\w^\tau_k) \right\|^2 \le \sigma_k^2$ and $\EB \left\| \nabla F_k(\w^\tau_k,\xi^\tau_k) \right\|^2  \le G^2$.
\end{assumption}

\subsection{Convergence Analysis}\label{sec:convergence}
Before formalizing the convergence theorem of FedSkip, we predefine $T$ as the total updates performed on local models and $\Gamma = F^* -\sum_{k=1}^{N}p_k F_k^*$ is a measure of the non-IID degree~\cite{convergence}, where $F^*$
and $F_k^*$ are the minimum values of $F$
and $F_k$, respectively. Then, based on the above assumptions, we have the following theorem (see the proof in the Appendix).
\begin{theorem}
        \label{thm:fedskip}
        Assume Assumptions~\ref{asm:smooth_and_strong_cvx} to \ref{asm:sgd_var_and_norm} hold and $L, \mu, \sigma_k, G$ are defined therein. The learning rate $\eta_\tau$ is non-increasing, non-negative and satisfies $\eta_\tau\leq\frac{1}{4L}$.
        Then, FedSkip with {the full device participation} satisfies
        \begin{equation}
        \label{eq:bound_K=N}
        \EB \left\|\overline{\w}^{T+1}-\w^{\star}\right\|^{2} \le \frac{32}{\mu}(\w^{0}-\w^{\star})\exp{\left[-\frac{\mu T}{2}\right]}+\frac{36C}{\mu^2 T},
        \end{equation}
where
\begin{equation}
C = \sum_{k=1}^N p_k^2 \sigma_k^2 + 6L \Gamma + 8  \Delta E^2G^2. \nonumber
\end{equation}
\end{theorem}
Compared to the convergence analysis of FedAvg~\cite{noniid1}, Theorem~\ref{thm:fedskip} introduces two distinct changes. On one hand, it is possible to acquire a lower variance of stochastic gradients $\sigma_k^2$ and a smaller non-IID degree $\Gamma$, since federated skip aggregation allows the local models to access data of more clients for a stable estimation. On the other hand, it does not perform federated averaging as frequently as FedAvg due to the skip operation, which introduces a scalar $\Delta$ on the local training iteration number $E$. Although the two-fold effect together cannot guarantee an absolutely smaller $C$, 
the performance in Section~\ref{sec:accuracy} and Section~\ref{sec:larger} implicitly confirms that we actually achieve a better convergence rate.

\subsection{Complexity Analysis}\label{sec:complexity}
\red{On the client side, FedSkip does not introduce any computation or storing cost compared with FedAvg, while methods including MOON, FedProx, FedDyn and SCAFFOLD require the extra calculation and space for previous models or global models, especially MOON needs to compute the representations of the input by multiple times.} On the server side, FedSkip is the same as FedAvg both in terms of time and space consumption in the communication round $t\notin I_{skip}$. However, in the communication round $t\in I_{skip}$, FedSkip is quite different from FedAvg. Specifically, in this case, FedSkip saves the computation time of aggregation but consumes the additional time to shuffle the received local models for scattering. Thus, their relation on time complexity is
$O_\text{FedSkip}=O_\text{FedAvg}+\frac{\Delta-1}{\Delta}T(S-A)$
where $O_\text{FedSkip}$ and $O_\text{FedAvg}$ represent the time complexity of FedSkip and FedAvg. S and A respectively denote the shuffle time and the aggregation time in one round, and T is the maximal round. \red{Although only FedNova of all baseline methods needs to additionally compute aggregation weights, the shuffle operation of FedSkip does not cause much computing overhead since it is random and does not need any tricks. In terms of the space complexity, since all approaches need to receive and distribute models, they can be approximately equal}.
\section{Experiments}\label{sec:experiment}
\subsection{Experimental Setup}\label{sec:setup}
\subsubsection{Dataset}
\begin{figure*}[!t]
    \centering
    \subfigure[IID ($\beta=\infty$)]{
    \centering
    \label{iid}
    \includegraphics[width=5.2cm]{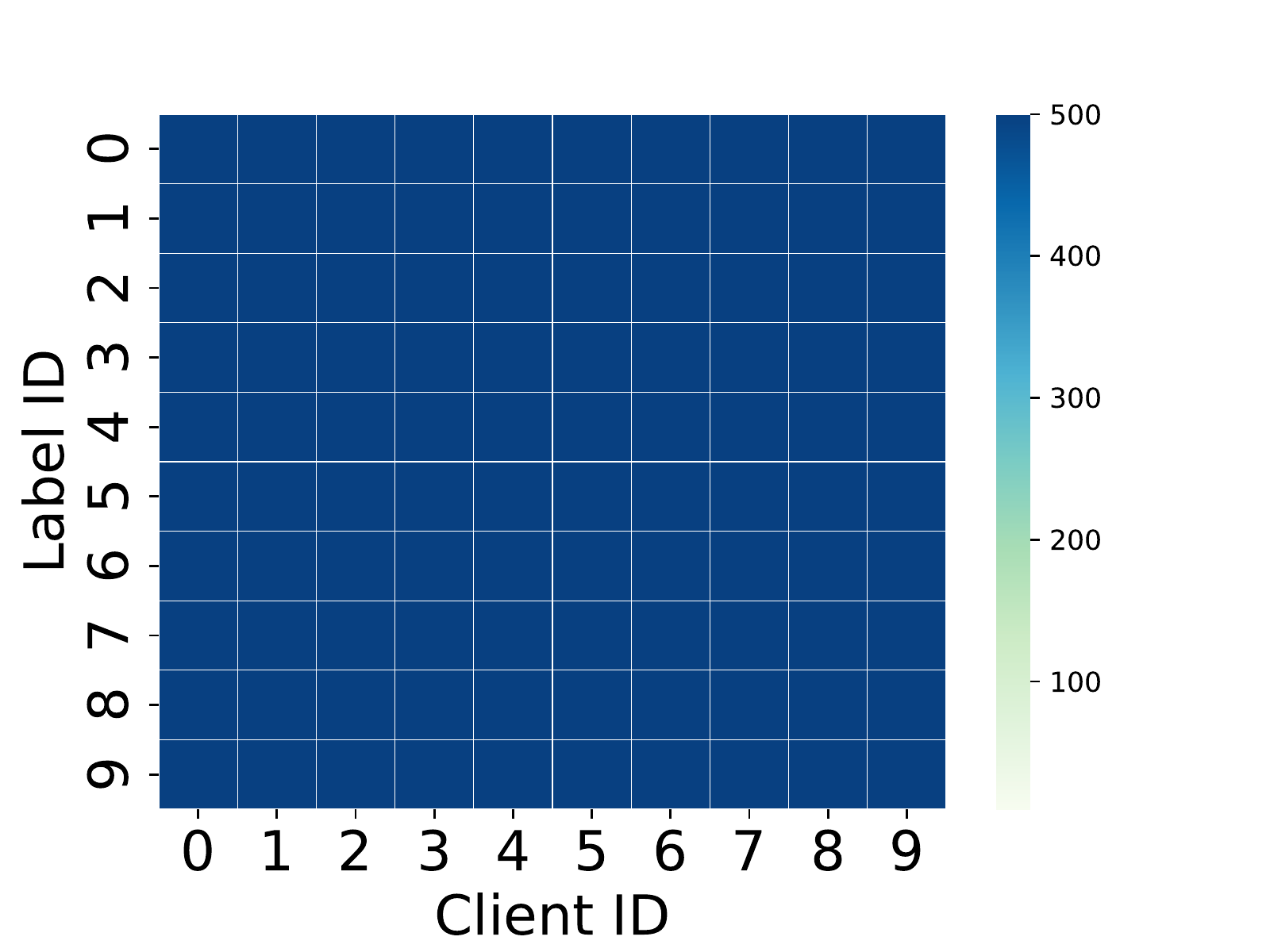}}
    \subfigure[Non-IID ($\beta=1$)]{
    \centering
    \label{beta1}
    \includegraphics[width=5.2cm]{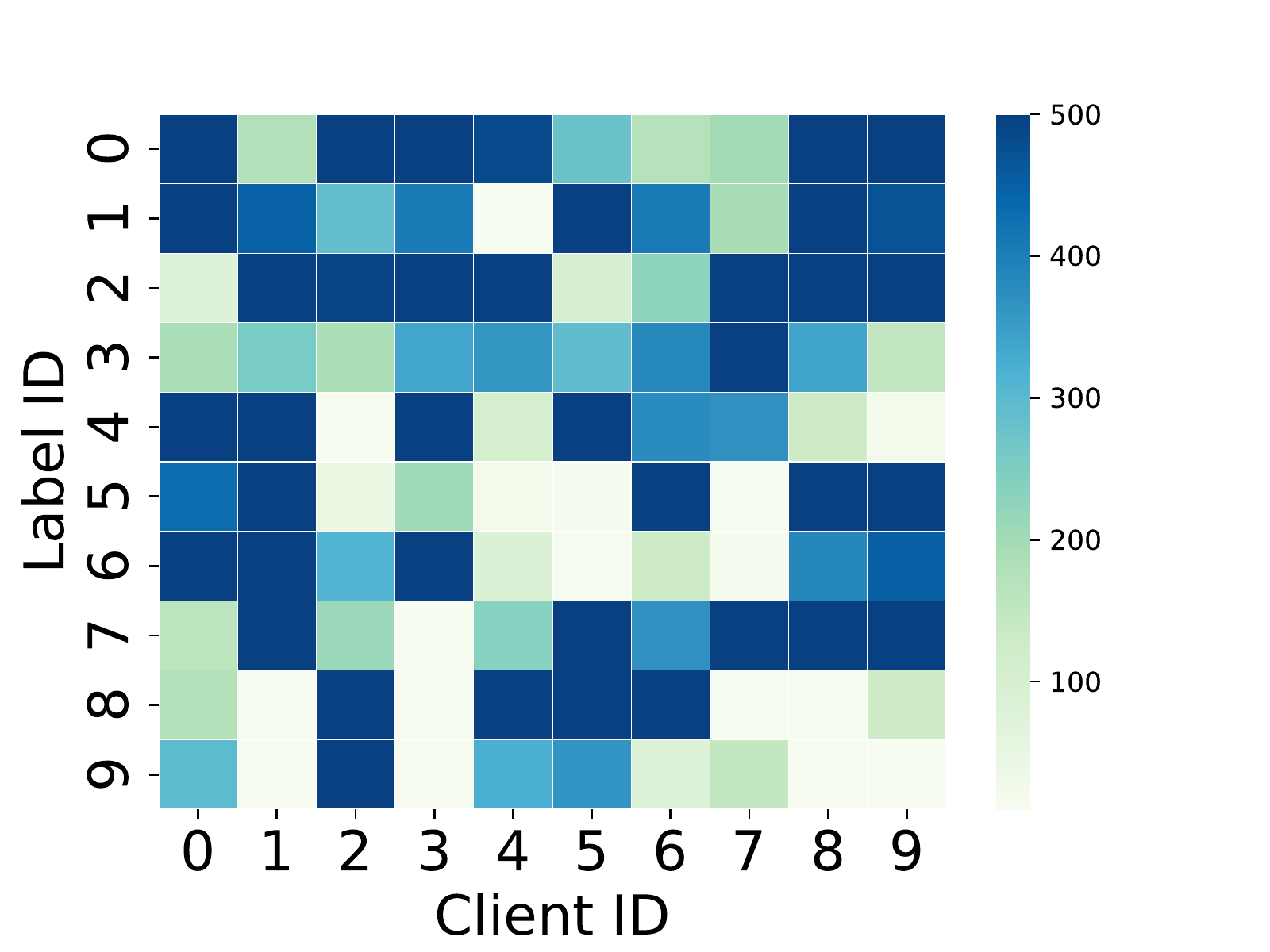}}
    \subfigure[Non-IID ($\beta=0.5$)]{
    \centering
    \label{beta05}
    \includegraphics[width=5.2cm]{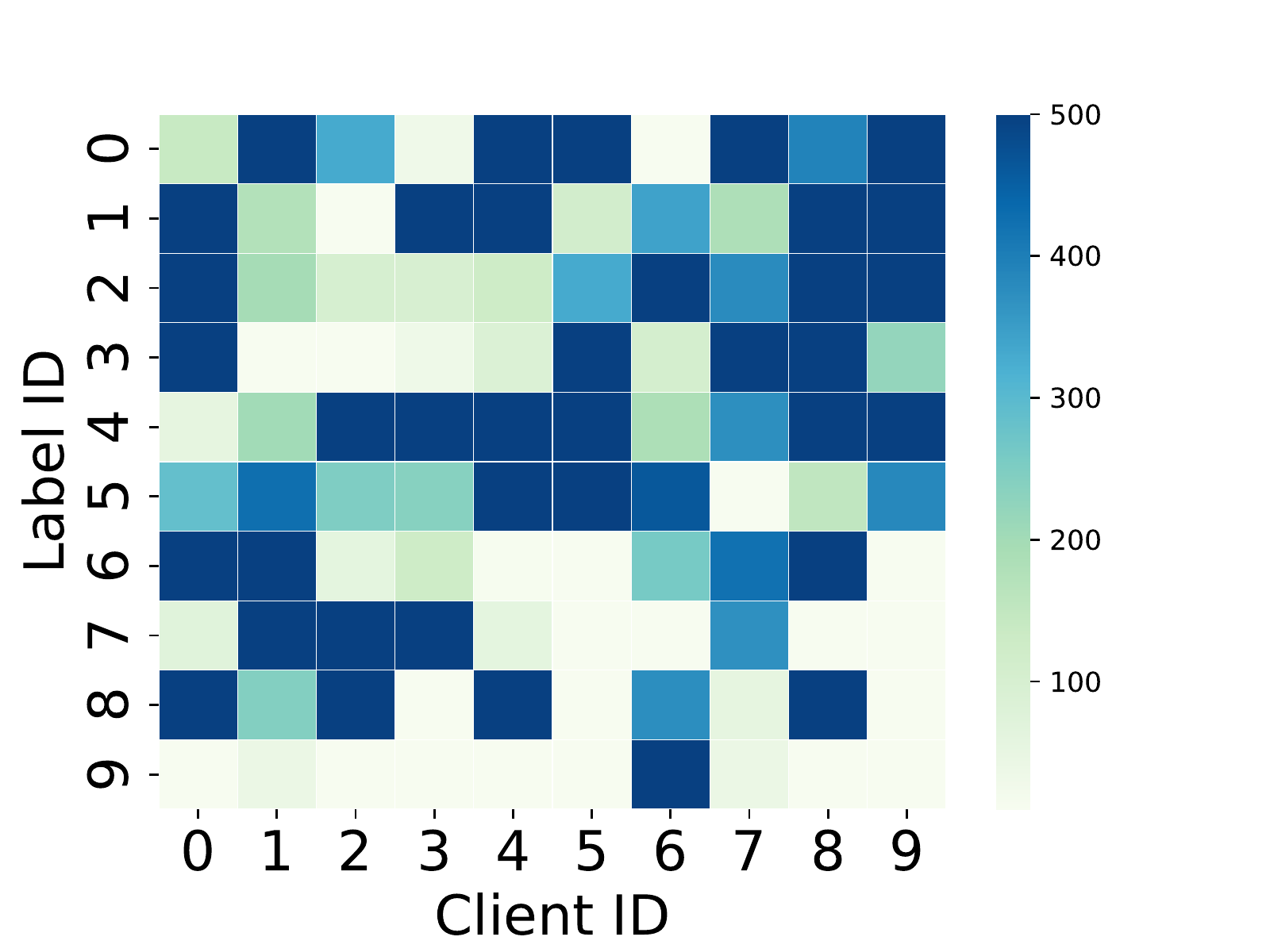}}
    \caption{Visualization of the data distribution in local clients on CIFAR-10~(P10) based on Dirichlet sampling under IID~($\beta=\infty$), $\beta=1$ and $\beta=0.5$ (\textit{i.e.,} non-IID). 
    Each grid reflects the number of samples of a client class. The darker the color is, the more samples it is allocated to the client.}\label{fig:distribution}
\end{figure*}

We leverage the popular CIFAR-10, CIFAR-100~\cite{cifar}, SYNTHETIC~\cite{synthetic,femnist1}, SHAKESPEARE~\cite{shakespeare,fedavg} and FEMNIST~\cite{femnist1,femnist2} in federated learning to conduct the experiments. Following~\cite{noniid2,dili2,dili3,moon,noniid2}, the Dirichlet sampling $p_{k,j}\sim Dir_N(\beta)$ is applied to generate the statistical heterogeneous partitions of CIFAR-10 and CIFAR-100, where $p_{k,j}$ denotes the probability of allocating a sample of label $k$ to client $j$. In Fig.~\ref{fig:distribution}, we divide CIFAR-10 into 10 clients and illustrate the statistics of the partitions under different $\beta$. Compared to $\beta=1$ in Fig.~\ref{beta1}, when $\beta$ becomes larger~($\beta=\infty$) or smaller~($\beta=0.5$) as shown in Fig.~\ref{iid} and \ref{beta05}, the data distributions of local clients are more even or more diverse. Based on the above strategy, CIFAR-10, CIFAR-100 are split into 10 clients, termed as P10, and all clients participate in the training in each communication round. Moreover, to compare the performance in the larger-scale of clients with the partial participation, we split them into 50 and 100 clients, denoted as P50 and P100 respectively. In each round, we randomly sample $20\%$ of clients to simulate partial participation. \red{Regarding the other three datasets, FEMNIST is a complex 62-class handwriting dataset and each client is one of the writers. SHAKESPEARE is the task of next-character prediction and each client represents a speaking role in the plays. SYNTHETIC is the task of five classes classification with one cluster and 60 features. We generate the three datasets by two steps: 1) generate a small-sized dataset of FEMNIST and full-sized datasets of SYNTHETIC and SHAKESPEARE with help of LEAF~\cite{femnist1}~(a benchmark
for federated settings) and then 2) remove clients with less than 64 training samples~(batch size of local training). After the two steps, we have 212, 180 and 660 clients of SYNTHETIC, FEMNIST and SHAKESPEARE respectively. In each round, we randomly sample 10 clients. Please refer to LEAF\cite{femnist1} for more detailed information of the first step.} 
\subsubsection{Implementation}
we compare FedSkip with FedAvg~\cite{fedavg} and several recent state-of-art methods including FedProx~\cite{fedprox}, SCAFFOLD~\cite{scaffold}, MOON~\cite{moon}, FedNova~\cite{fednova} and FedDyn~\cite{fedyn}. \red{To make a fair and comprehensive comparison, we utilize multiple model structures based on several previous studies. Concretely, for CIFAR-10 and FEMNIST, we use a simple CNN as backbone while for CIFAR-100, ResNet-50~\cite{resnet} is used as backbone referring to MOON~\cite{moon} and 
the experimental study~\cite{noniid2}. For SHAKESPEARE, we adopt two-layer LSTM classifier containing 100 hidden units with an 8D embedding layer according to FedProx~\cite{fedprox} and LEAF~\cite{femnist1}. The model of SYNTHETIC is the same as LEAF: a perceptron with sigmoid activations. The optimizer of all datasets and methods is SGD with a learning rate 0.01, the weight decay $1e-5$ and momentum 0.9. The batch size is set to 64 and the local updates are set to 10 epochs for all approaches. All methods are implemented in PyTorch~\cite{pytorch}}.


\begin{table}[!t]
    \centering
    \caption{\rm{Global test accuracy of FedSkip and a range of state-of-the-art approaches on CIFAR-10 and CIFAR100 under IID and non-IID~($\beta=1$ and $\beta=0.5$) partitions. P10 is denoted the total number of clients and $\Delta$ from FedSkip~($\Delta$) denotes the number of skip between two near aggregations.}}
    \setlength{\tabcolsep}{2pt}
    \small
    \begin{tabular}{llllllllll}
        \toprule
        \multirow{2}{*}{Method}& 
        \multicolumn{3}{c}{CIFAR-10~(P10)}&
        \multicolumn{3}{c}{CIFAR-100~(P10)}& & 
        \\
        \cmidrule(r){2-4}\cmidrule(r){5-7}
        \multicolumn{1}{c}{} & IID& $\beta=1$&$\beta=0.5$& IID& $\beta=1$&$\beta=0.5$\\
        \midrule
        FedAvg   &$0.7285$ &$0.7021$ &$0.6883$ &$0.6711$ &$0.6579$ &$0.6563$\\
        FedProx  &$0.7293$ &$0.7132$ &$0.6680$ &$0.6695$ &$0.6595$ &$0.6537$\\
        SCAFFOLD &$\underline{0.7432}$ &$0.7113$&$\underline{0.7053}$ &$\underline{0.7127}$&$\underline{0.7066}$&$\underline{0.6991}$\\
        MOON&$0.7307$ &$0.7162$ &$0.6932$ &$0.6803$ &$0.6798$ &$0.6781$ \\
        FedNova  &$0.7354$ &$0.7200$ &$0.6703$ &$0.6721$ &$0.6601$ &$0.6543$ \\
        FedDyn  &$0.7430$ &$\underline{0.7328}$ &$0.6965$ &$0.6783$ &$0.6750$ &$0.6717$ \\
        \midrule
        FedSkip(3) &$\textbf{0.7453}$ &$0.7322$ &$0.7146$ &$0.6908$ &$0.6895$ &$0.6805$ \\
        FedSkip(5) &$0.7374$ &$0.7357$ &$\textbf{0.7177}$ &$0.7106$ &$0.7007$ &$0.6971$ \\
        FedSkip(7) &$0.7374$ &$0.7355$ &$0.7155$ &$0.7287$ &$\textbf{0.7146}$ &$\textbf{0.7174}$ \\
        FedSkip(10)&$0.7384$ &$\textbf{0.7368}$ &$0.7136$ &$\textbf{0.7317}$ &$0.7029$ &$0.7015$ \\
        \bottomrule
    \end{tabular}
    \label{tab:accuracy1}
\end{table}

\begin{table*}[!t]
    \centering
    \caption{\rm{Accuracy of FedSkip and baselines on CIFAR-10, CIFAR-100, SYNTHETIC, FEMNIST and SHAKESPEARE. CIFAR-10 and CIFAR-100 are divided into 50 and 100 clients under non-IID~($\beta=0.5$) partitions. FedSkip-$\Delta$ means that we set FedSkip with the period $\Delta$ in \eqref{eq:skip}. Besides, the best results of FedSkip are marked in bold and the best results of baselines are underlined, which are along with the improvement in subscript relative to FedAvg. We use $*$ to denote the convergence failure of SCAFFOLD and the failure of applying MOON on SHAKESPEARE due to the LSTM backbone.}}
    \setlength{\tabcolsep}{2pt}
    \small

    \begin{tabular}{llllllllllll}
        \toprule
        \multirow{2}{*}{Method}& 
        \multicolumn{1}{c}{SYNTHETIC}& 
        \multicolumn{1}{c}{FEMNIST}& 
        \multicolumn{1}{c}{SHAKESPEARE}& 
        \multicolumn{2}{c}{CIFAR-10~($\beta=0.5$)}& 
        \multicolumn{2}{c}{CIFAR-100~($\beta=0.5$)}& 
        \\
        \cmidrule(r){2-2}\cmidrule(r){3-3}\cmidrule(r){4-4}\cmidrule(r){5-6}\cmidrule(r){7-8}
        \multicolumn{1}{c}{} & (212,10) & (180,10) & (660,10) & (50,10) & (100,20) & (50,10)& (100,20)\\
        \midrule
        FedAvg   &$0.9412$ &$0.6347$ &$0.5098$ &$0.6821$ &$0.6669$ &$0.5997$&$0.5956$\\
        FedProx  &$0.9394$ &$0.6407$ &$0.5081$ &$0.6832$ &$0.6618$ &$0.6025$&$0.5825$\\
        SCAFFOLD &$*$ &$*$&$*$ &$*$&$*$&$*$&$*$\\
        MOON     &$0.9417$&$0.6407$ &$*$ &$\underline{0.6970}_{1.5\%\uparrow}$ &$0.6710$ &$0.6138$ &$0.6165$\\
        FedNova  &${\underline{0.9472}}_{0.6\%\uparrow}$ &$0.6407$ &${\underline{0.5216}}_{1.2\%\uparrow}$ &$0.6880$ &$0.6674$ &${\underline{0.6660}}_{6.6\%\uparrow}$&${\underline{0.6615}}_{6.6\%\uparrow}$\\
        FedDyn  &$0.9459$ &${\underline{0.6527}}_{1.8\%\uparrow}$ &$0.5181$ &$0.6904$ &$\underline{0.6770}_{1.0\%\uparrow}$ &$0.6128$&$0.6006$\\
        \midrule
        FedSkip-3 &$\textbf{0.9481}_{0.7\%\uparrow}$ &$0.6467$ &$0.5217$ &$0.6903$ &$0.6805$ &$0.6536$&$0.6615$\\
        FedSkip-5 &$0.9394$ &$0.6586$ &$0.5155$ &$0.6911$ &$0.6765$ &$0.6812$&$0.6859$\\
        FedSkip-7 &$0.9410$ &$\textbf{0.6766}_{4.2\%\uparrow}$ &$\textbf{0.5232}_{1.3\%\uparrow}$ &$0.6852$ &$\textbf{0.6864}_{2.0\%\uparrow}$&$0.6962$ &$0.6971$\\
        FedSkip-10&$0.9450$&$0.6766$ &$0.5220$ &${\textbf{0.6982}}_{1.6\%\uparrow}$ &$0.6790$ &$\textbf{0.6978}_{9.8\%\uparrow}$ &$\textbf{0.7002}_{10.5\%\uparrow}$\\
        \bottomrule
    \end{tabular}
    \label{tab:accuracy2}
\end{table*}
\subsection{Performance under IID \textit{vs.} Non-IID}\label{sec:accuracy}
Following the aforementioned basic settings, we split CIFAR-10 and CIFAR-100 into 10 clients under both IID~($\beta=\infty$) and non-IID ($\beta=1$ and $\beta=0.5$) settings and compare FedSkip with baselines. Note that, FedProx, MOON and FedDyn have a hyper-parameter $\mu$ to respectively control the weight of the proximal term, the contrastive loss and the dynamic regularization. \red{We follow~\cite{moon} that for CIFAR-10 and CIFAR-100, $\mu$ of FedProx are chosen as 0.01 and $\mu$ of MOON are chosen as 5 and 1 respectively. For FedDyn, we search the best parameter 0.001 and 0.0001 from \{0.01, 0.001, 0.0001, 0.00001\}. For all approaches, the number of communication rounds is set to 200. }

Table~\ref{tab:accuracy1} shows the Top-1 classification performance of all methods on the test set under the IID scenario ($\beta=\infty$) and the non-IID scenario~($\beta=1$ and $\beta=0.5$). From the results, we can find that the degradation happens when the data becomes heterogeneous, \textit{i.e.,} from IID to non-IID. \red{And the more heterogeneous the setting is, the worse the performance drops. Among all approaches  under the IID or non-IID settings, FedSkip achieves the best performance, which confirms the benefit of routing local models to access more data.}

\begin{figure*}[!ht]
    \centering
    \subfigure[CIFAR-100~(P10)]{
    \centering
    \label{cifar100beta05p10}
    \includegraphics[width=5.2cm]{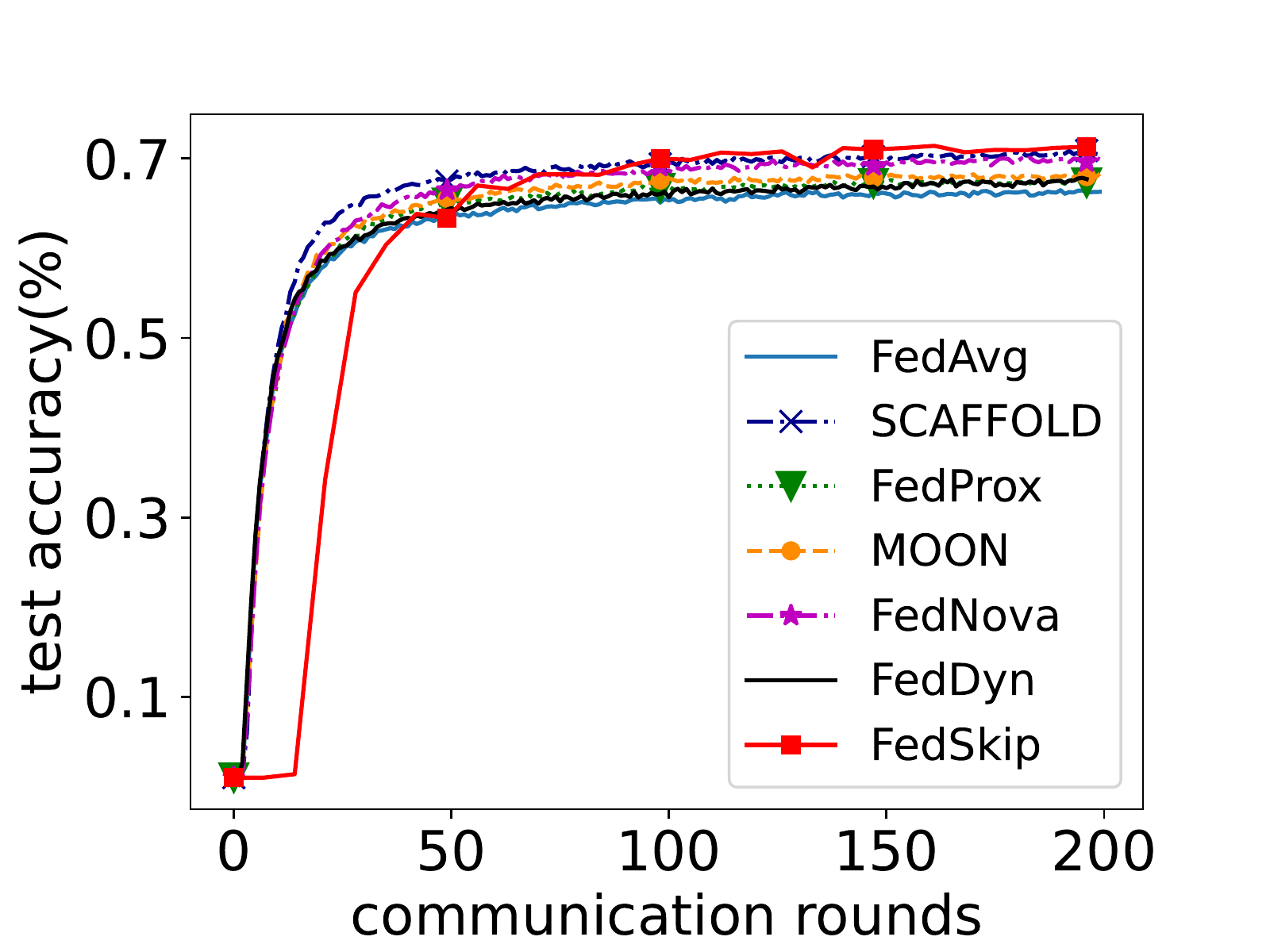}}
    \hspace{1em}
    \subfigure[CIFAR-100~(P50)]{
    \centering
    \label{cifar100beta05p50}
    \includegraphics[width=5.2cm]{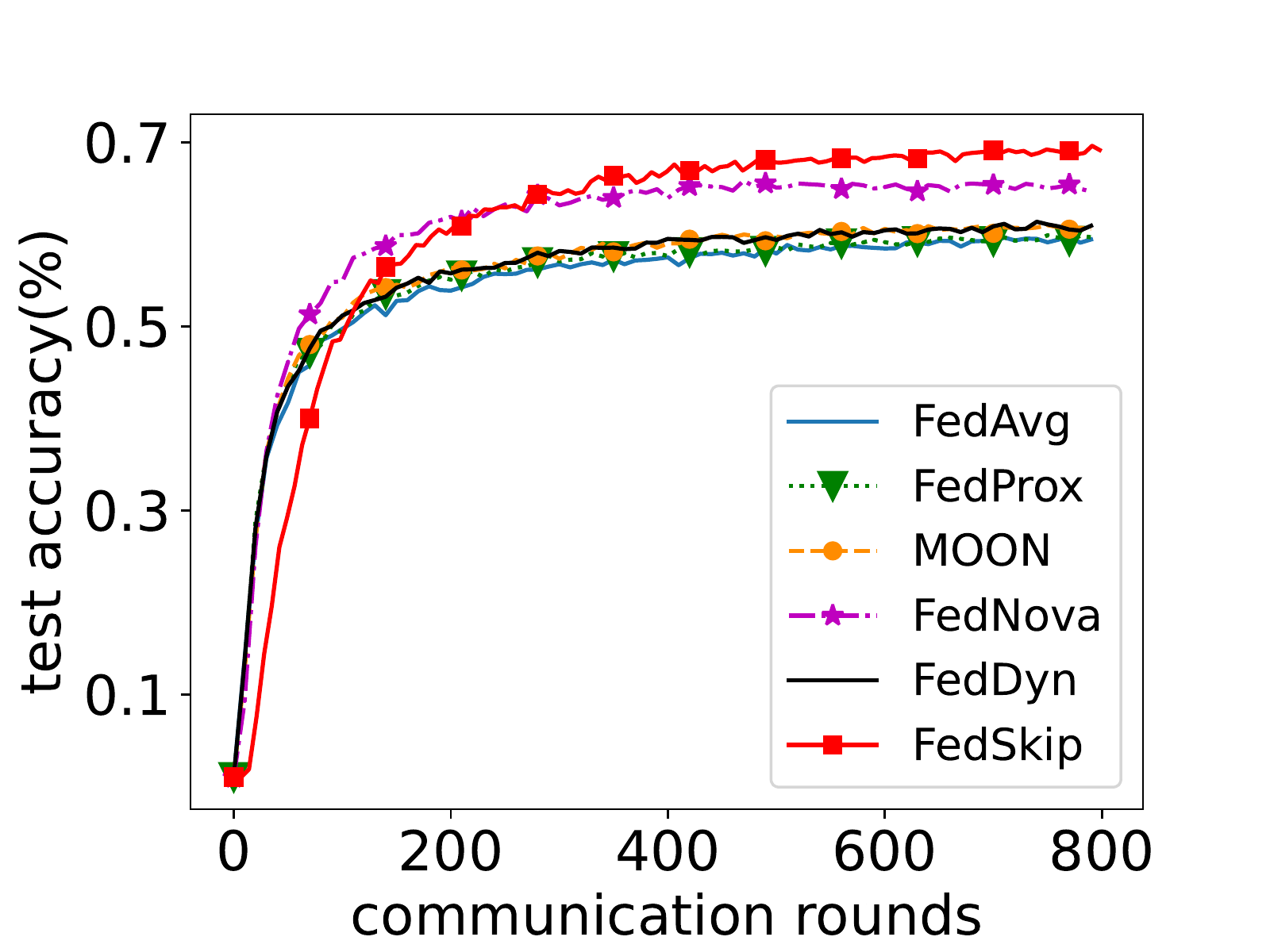}}
    \hspace{1em}
    \subfigure[CIFAR-100~(P100)]{
    \centering
    \label{cifar100beta05p100}
    \includegraphics[width=5.2cm]{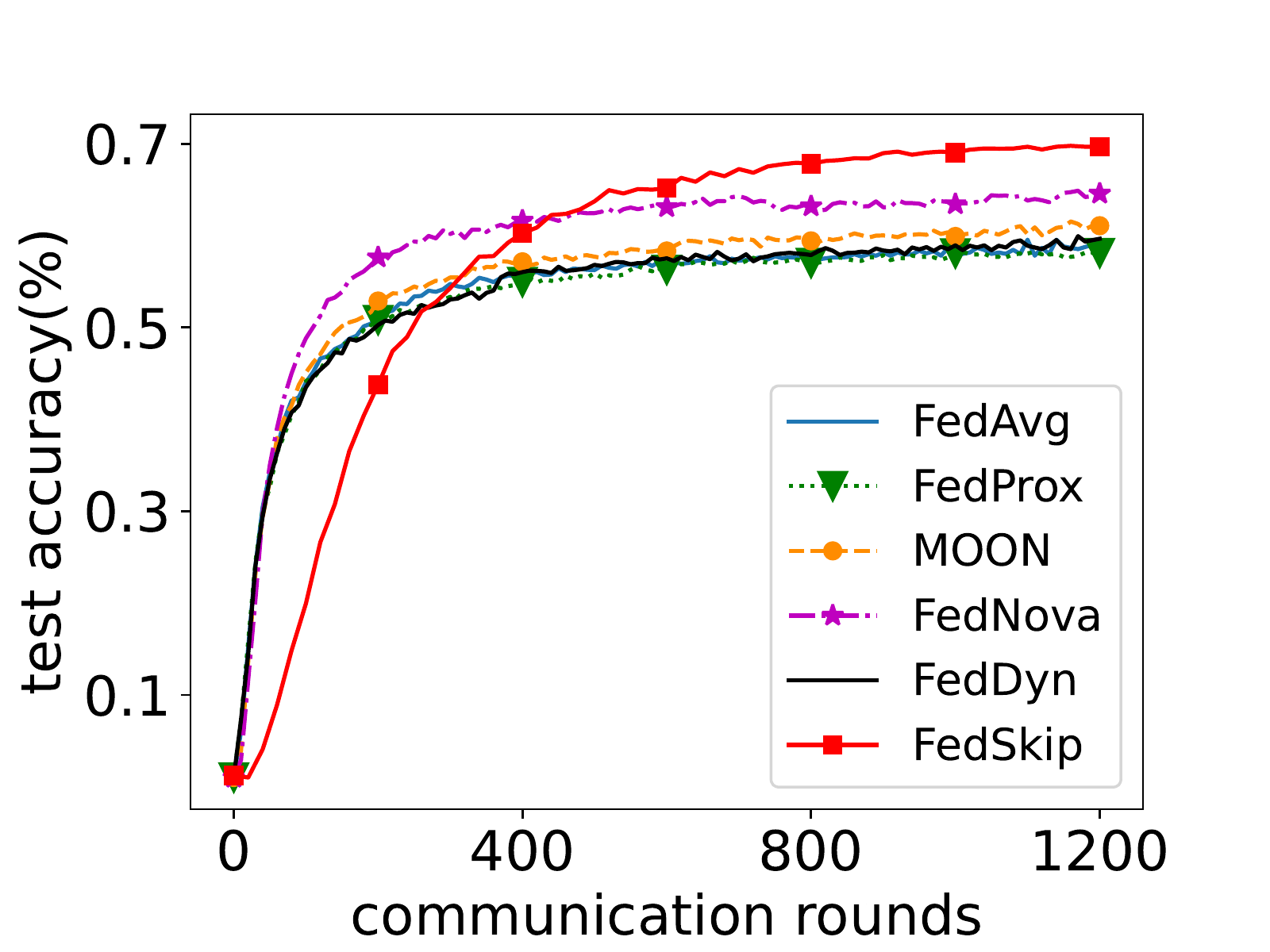}}
    \subfigure[SYNTHETIC]{
    \centering
    \label{synthetic}
    \includegraphics[width=5.2cm]{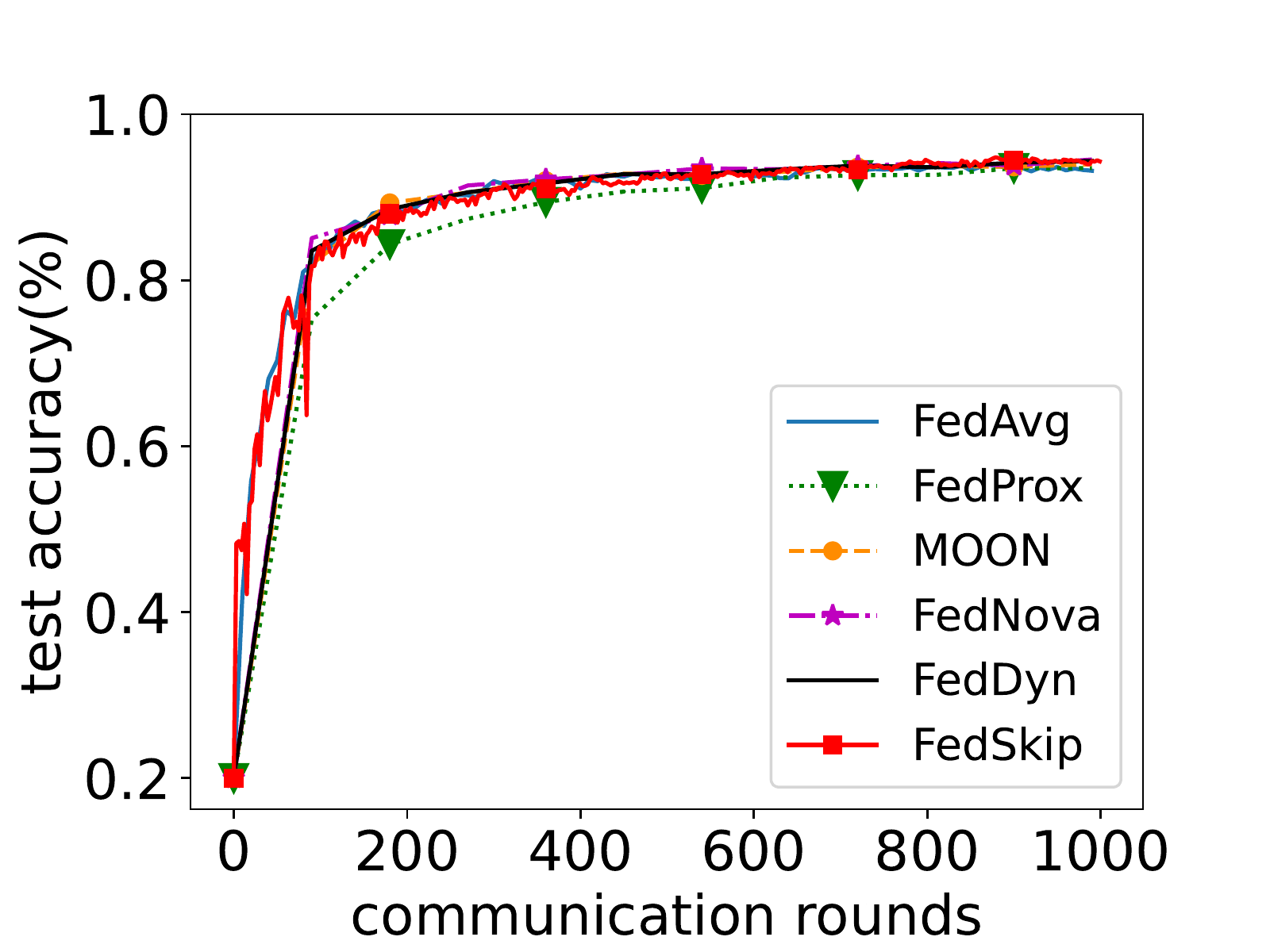}}
    \hspace{1em}
    \subfigure[SHAKESPEARE]{
    \centering
    \label{SHAKESPEARE}
    \includegraphics[width=5.2cm]{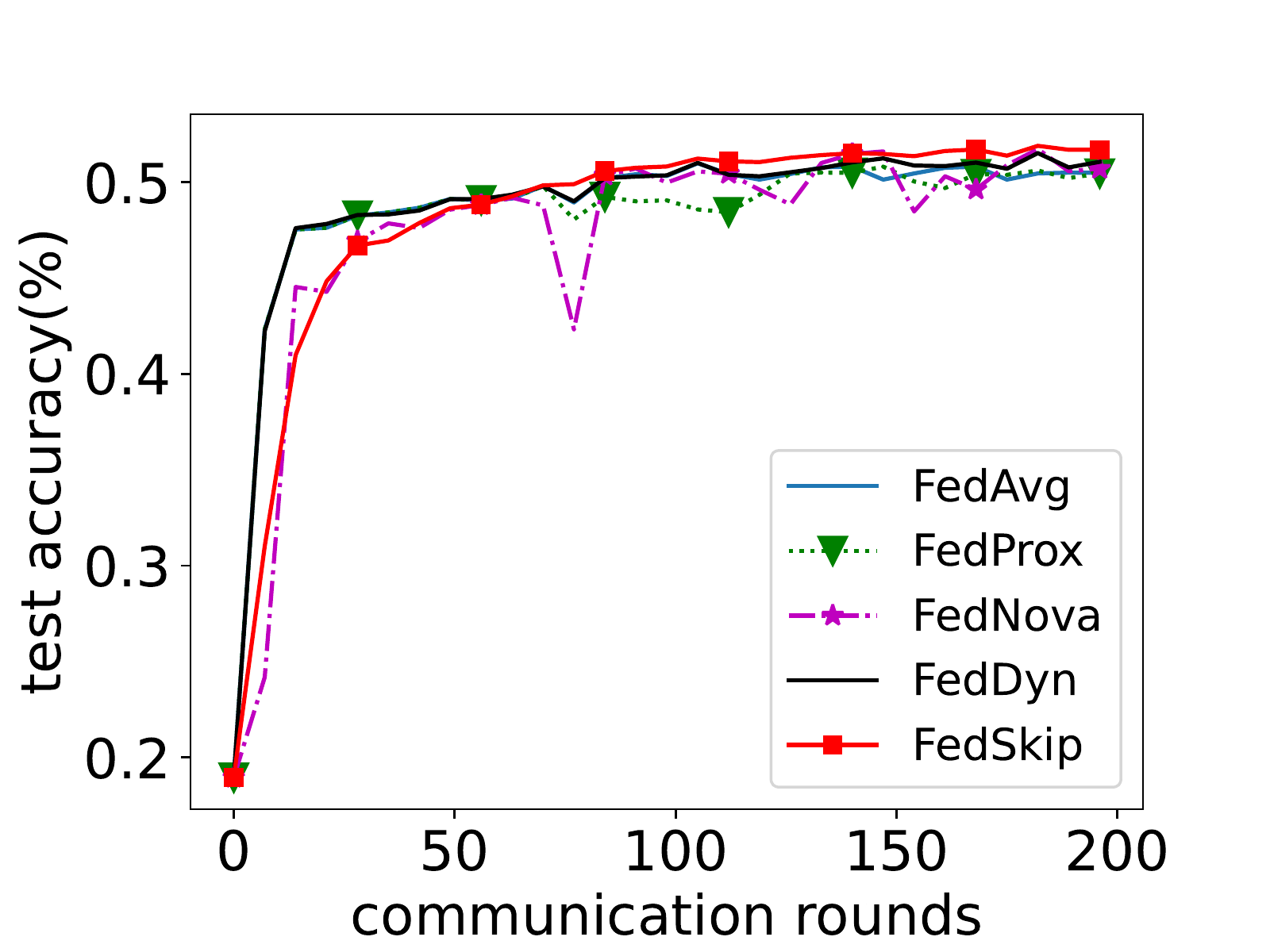}}
    \hspace{1em}
    \subfigure[FEMNIST]{
    \centering
    \label{femnist}
    \includegraphics[width=5.2cm]{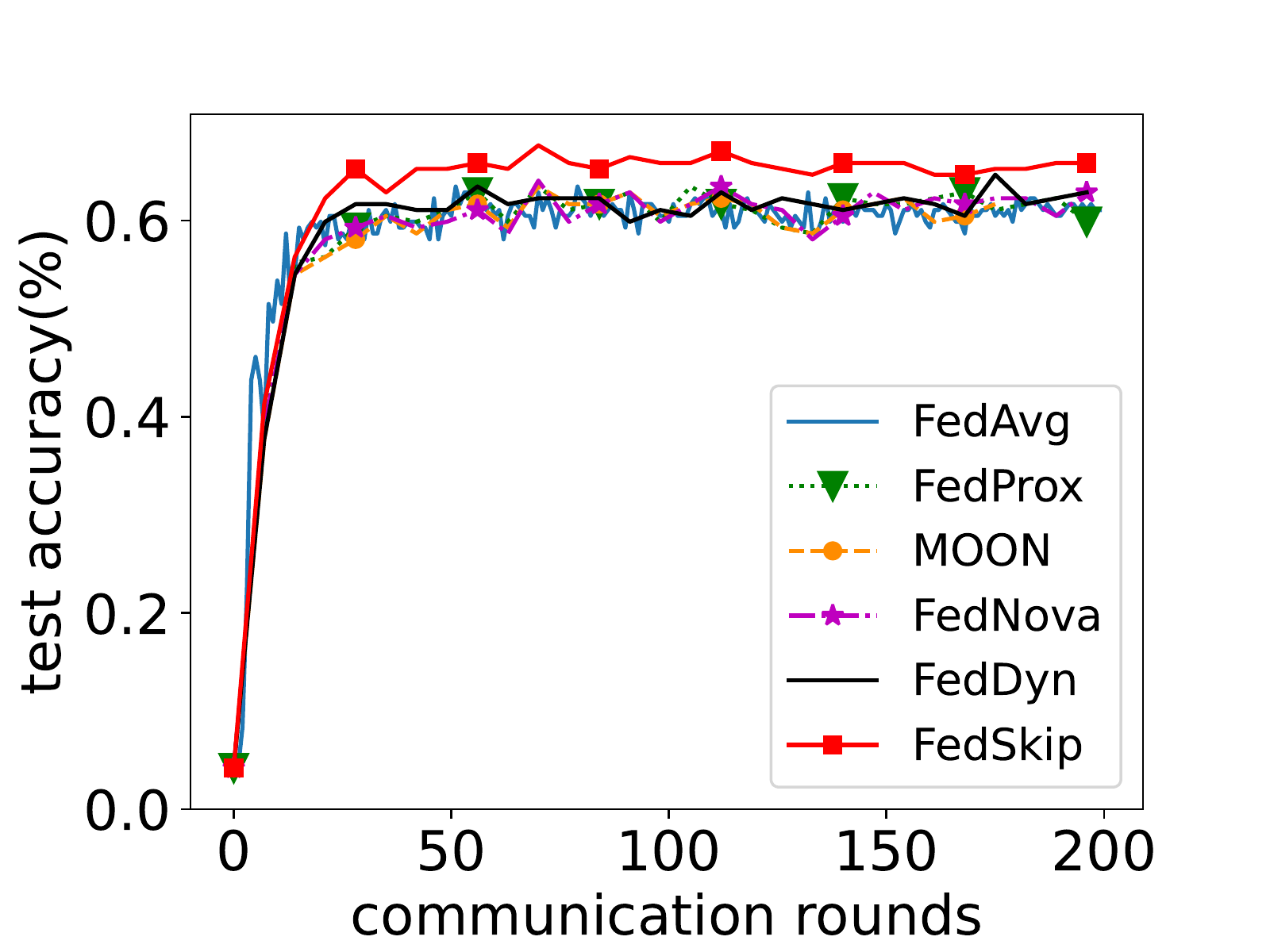}}
    \caption{The performance of FedSkip and baselines during the training process on  CIFAR-100, SYNTHETIC FEMNIST and SHAKESPEARE. The accuracy of FedSkip is lower in the first few rounds, but outperforms other methods finally with the sufficient federated skip aggregation and federated averaging.}\label{fig:procedure}
\end{figure*}
\subsection{Performance under Non-IID with Larger-Scale Clients, Real-world Scenarios and Partial Participation}\label{sec:larger}
In the previous section, we validate the FedSkip under 10 local clients, which is relatively small-scale. \red{In this part, we increase the client numbers of CIFAR-10 and CIFAR-100 and add one synthetic dataset and two real-world datasets, namely SYNTHETIC, FEMNIST and SHAKESPEARE. CIFAR-10 and CIFAR-100 are divided into 50 (P50) and 100 (P100) clients under non-IID~($\beta=0.5$) partitions and only $20\%$ of clients are randomly sampled to participate in the training in each communication round. SYNTHETIC, FEMNIST and SHAKESPEARE have 212, 180 and 660 local clients and 10 of clients are randomly selected in each round. To guarantee the model to converge, we set the number of total communication rounds to 200 for FEMNIST, 200 for SHAKESPEARE, 1000 for SYNTHETIC, 800 for CIFAR-10~(P50, P100) and CIFAR-100~(P50), 1200 for CIFAR-100~(P100). Hyper-parameter $\mu$ of MOON and FedProx is set to 1 and 0.001 for all datasets  and $\mu$ of FedDyn is set to 0.001, 0.0001 and 0.0001 for SYNTHETIC, FEMNIST and SHAKESPEARE respectively  tuning from $\{0.00001, 0.0001, 0.001, 0.01, 0.1, 1, 10\}$.} All other settings follow Section~\ref{sec:setup} and Section~\ref{sec:accuracy}.


\red{Table \ref{tab:accuracy2} reports the test accuracy of all methods in the larger number of clients, partially participation and real world heterogeneity. In the experiments, SCAFFOLD fails when the number of clients increases to a large scale and MOON also fails to apply on SHAKESPEARE due to the LSTM backbone. Thus, we denote SCAFFOLD of all results and MOON of SHAKESPEARE as *.} From the results, our approach FedSkip significantly outperforms all other approaches with much higher \red{accuracy on all datasets and partition strategies, especially $4.2\%$ improvement \textit{vs.} $1.2\%$ of FedDyn  for FEMNIST, $9.8\%$ improvement  \textit{vs.} $6.6\%$ of FedNova for CIFAR-100~(P50) and $10.5\%$ improvement \textit{vs.} $6.6\%$ of FedNova for CIFAR-100~(P100).} Besides, we can find that when the client number increases, the relative improvement of FedSkip becomes larger, which demonstrates the scalability of our method. In Fig.~\ref{fig:procedure}, we illustrate the test performance of each method in the whole training procedure. From the figure, FedSkip does not achieve the better performance in the first few rounds, but after the sufficient federated skip aggregation and federated averaging, it finally achieves the best results. This indicates although federated skip aggregation enables the client model to access more data in the local training, federated averaging for model aggregation is also critical in FedSkip.

\begin{table*}[!t]
    \centering
    \caption{\rm{Number of aggregation times, communication rounds and the speedup of communication when reaching the best accuracy of FedAvg in federated learning on CIFAR-100 under four partition strategies. We use $*$ to denote the convergence failure of SCAFFOLD. P10, P50 and P100 are used to denote the heterogeneous partition strategies as mentioned in Section~\ref{sec:setup} and $ \Delta$ from FedSkip-$\Delta$ denotes the skip period. Although FedSkip is not the best from the perspective of the communication cost, it is appropriately the second best to greatly reduce the communication cost of FedAvg.}}
    \setlength{\tabcolsep}{2pt}
    \small
    \begin{tabular}{llllllllllllll}
        \toprule
        \multirow{2}{*}{Method}& 
        \multicolumn{3}{c}{CIFAR-100~(P10, IID)}&\multicolumn{3}{c}{CIFAR-100~(P10)}& 
        \multicolumn{3}{c}{CIFAR-100~(P50)}&
        \multicolumn{3}{c}{CIFAR-100~(P100)}
        \\
        \cmidrule(r){2-4}\cmidrule(r){5-7}\cmidrule(r){8-10}\cmidrule(r){11-13}
        \multicolumn{1}{c}{} & Arrgre.& Commu. & Speedup& Arrgre.& Commu. & Speedup& Arrgre.& Commu. & Speedup& Arrgre.& Commu. & Speedup  \\
        \midrule
        FedAvg&$200$&$200$ &$1\times$&$200$&$200$ &$1\times$ &$800$&$800$  &$1\times$&$1200$&$1200$&$1\times$\\
        FedProx&$248$&$248$ &$0.81\times$&$114$&$114$  &$1.75\times$&$762$&$762$&$1.05\times$&$1321$&$1321$&$0.91\times$\\
        SCAFFOLD&$24$&$24$ &$\underline{8.33}\times$&$40$&$40$       &$\underline{5.00}\times$&$*$&$*$&$*$&$*$&$*$&$*$\\
        FedNova&$91$&$91$ &$2.20\times$&$48$&$48$       &$4.17\times$&$144$&$144$&$\underline{5.56}\times$&$187$&$187$&$\underline{6.42}\times$\\
        MOON&$41$&$41$ &$4.88\times$&$65$&$65$  &$3.08\times$&$450$&$450$&$1.78\times$&$651$&$651$&$1.84\times$\\
        FedDyn&$109$&$109$ &$1.83\times$&$71$&$71$  &$2.82\times$&$435$&$435$&$1.84\times$&$1104$&$1104$&$1.09\times$\\
        \midrule
        FedSkip-3&$11$&$33$ &$\textbf{6.06}\times$&$22$&$66$  &$1.52\times$&$77$&$231$&$3.46\times$&$153$&$459$&$2.61\times$\\
        FedSkip-5&$8$&$40$ &$5.00\times$&$12$&$60$   &$\textbf{3.33}\times$&$38$&$190$&$4.21\times$&$90$&$450$&$\textbf{2.67}\times$\\
        FedSkip-7&$6$&$42$ &$4.76\times$&$9$&$63$   &$3.17\times$&$27$&$189$&$\textbf{4.23}\times$&$70$&$490$&$2.45\times$\\
        FedSkip-10&$5$&$50$ &$4.00\times$&$8$&$80$   &$2.50\times$&$19$&$190$&$4.21\times$&$53$&$530$&$2.26\times$\\
        \bottomrule
    \end{tabular}
    \label{tab:commu}
\end{table*}
\begin{figure*}[!ht]
    \centering
    \subfigure[CIFAR-100~(P10)]{
    \centering
    \label{fig:abl_C100P10}
    \includegraphics[width=4cm]{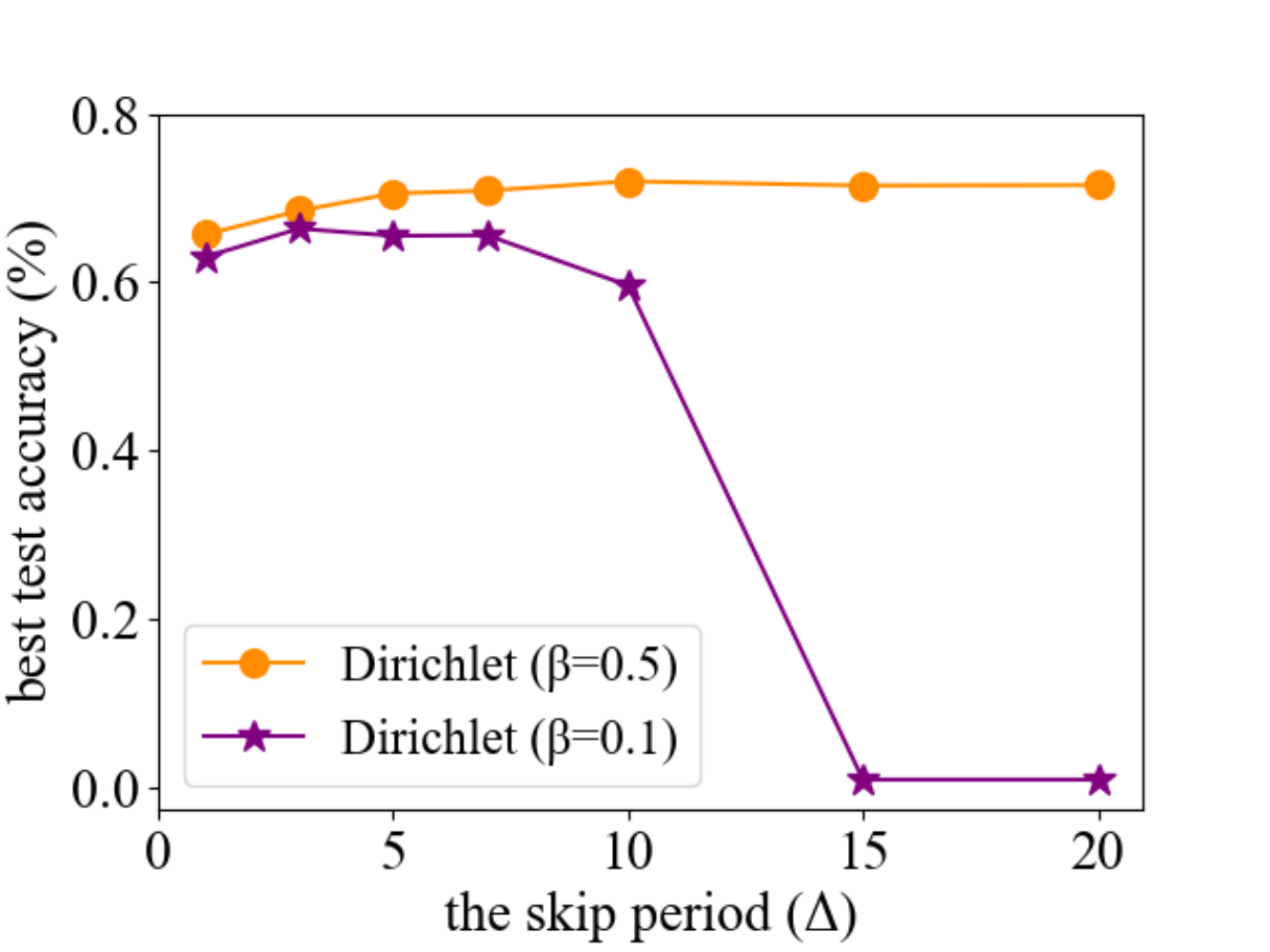}}
    \subfigure[CIFAR-100~(P50)]{
    \centering
    \label{fig:abl_C100P50}
    \includegraphics[width=4cm]{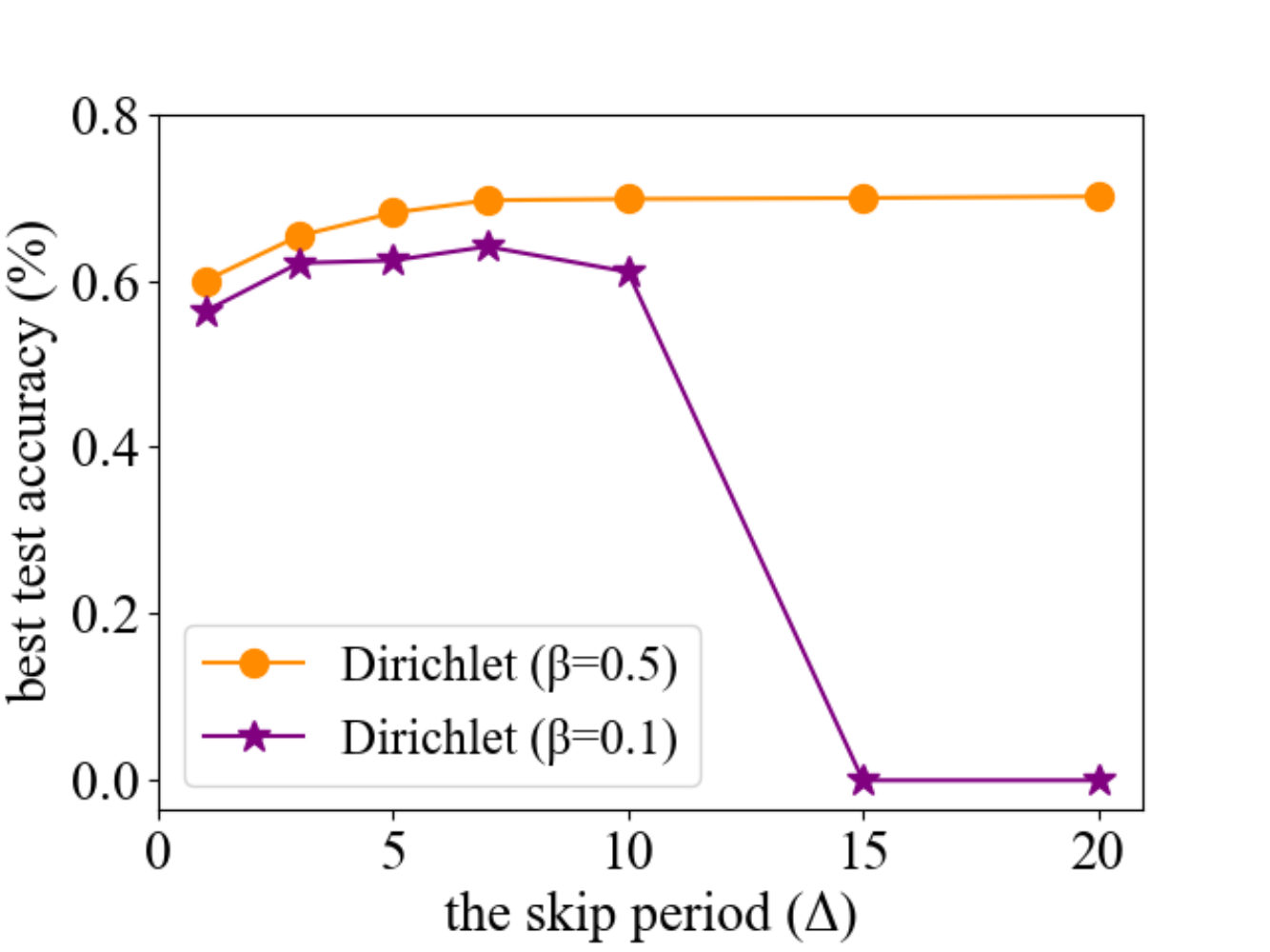}}
    \subfigure[CIFAR-10~($\beta=0.5$)]{
    \centering
    \label{fig:abl_c10n}
    \includegraphics[width=4cm]{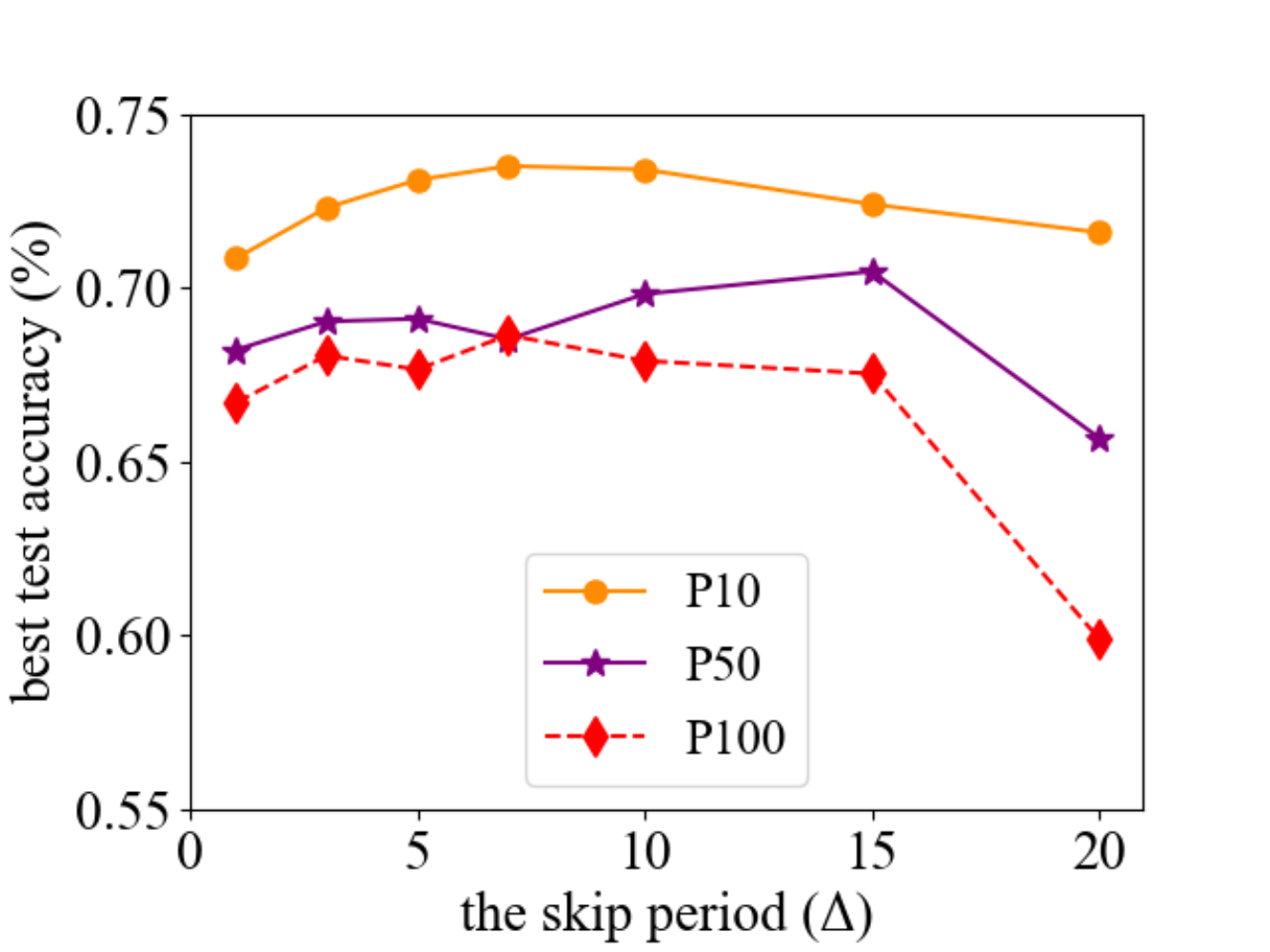}}
    \subfigure[CIFAR-100~($\beta=0.5$)]{
    \centering
    \label{fig:abl_c100n}
    \includegraphics[width=4cm]{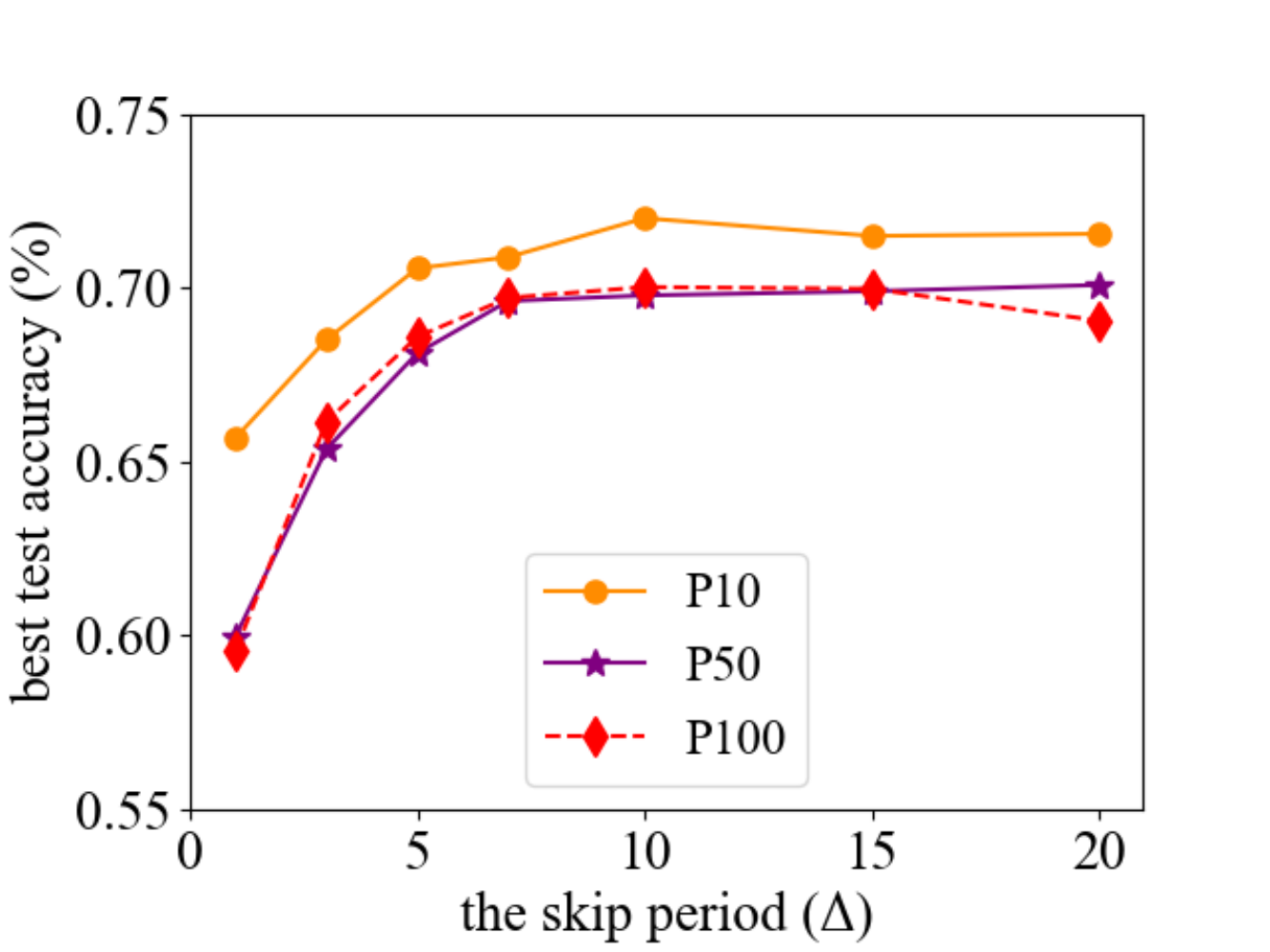}}
    \caption{The performance of FedSkip \textit{w.r.t} $\Delta$ under different settings. In (a) and (b), we plot $\Delta$ with different $\beta$ under P10 and P50 on CIFAR-100. In (c) and (d), we show $\Delta$ with different client numbers on both CIFAR-10 and CIFAR-100 ($\beta=0.5$).}\label{fig:ablation}
\end{figure*}

\subsection{On Communication Cost and Aggregation Efficiency}\label{sec:efficiency}
In addition to the comparison of the test accuracy, we count the communication rounds and aggregation times of all approaches when they reach the best accuracy of FedAvg on CIFAR-100 under different partitions. According to the results, FedSkip greatly reduces the communication cost of FedAvg, although it is generally the second best improvement among all approaches. 
Compared to the best cases, FedSkip requires slightly more communication rounds because it gives up the chances for the model aggregation when we make the model access data of more clients in the skip period $\Delta$. From the empirical observation in Fig.~\ref{fig:procedure}, the model aggregation is also critical and necessary to achieve faster convergence, especially in the early stage of FedSkip. However, in terms of the truly effective aggregation times that are computed by the communication rounds to divide the skip period, our method is the most efficient. \red{This indicates in the same aggregation stage, our method is more efficient and local models from FedSkip are in the better conditions compared with all other approaches since it is trained on more local data.}


\subsection{Ablation Study: The Sensitivity of Skipping Rounds}\label{sec:ablation}
In this section, we explore the impact of $\Delta$ on the performance of FedSkip with different clients (P10 and P50) and different statistical heterogeneity ($\beta=0.5$ and $0.1$). All experiments reported are under the same setting as in the previous sections. In Fig.~\ref{fig:abl_C100P10} and~\ref{fig:abl_C100P50}, we present the best test accuracy of FedSkip with different $\Delta$ and $\beta$ on CIFAR-100~(P10) and CIFAR-100~(P50). From the curves, we can see that the performance of the less heterogeneous data is relatively stable with different $\Delta$. However, in the more heterogeneous case ($\beta=0.1$), too large $\Delta$ induces the performance degeneration. This means in more heterogeneous environments, compared to only focusing on accessing more data, the more frequent model aggregation combined with federated skip aggregation is also important to boost the final performance. In Fig.~\ref{fig:abl_c10n} and~\ref{fig:abl_c100n}, we show the best test accuracy of FedSkip with different $\Delta$ and local clients (P10, P50, P100) on CIFAR-10 and CIFAR-100 ($\beta=0.5$). From the figures, we can find that different datasets demonstrate different preferences on $\Delta$ with different local clients. On CIFAR-10, it exhibits the decreasing performance with larger $\Delta$ while on CIFAR-100, it is still stable even when $\Delta=15$ and $20$.

\section{Conclusion}\label{sec:conclusion}
In this work, we are motivated by an observation on the client drift of federated learning under different heterogeneous scenarios and propose a new method FedSkip to combat the statistical heterogeneity. The intuition behind FedSkip is to make the local models see more clients' data without knowing their identities, so that the optima of client models are improved. Different from the previous method in the optimization perspective, FedSkip is totally data-driven without requiring the auxiliary terms to calibrate the learning bias or the objective inconsistency, but just with an effective federated skip aggregation (shuffle $\&$ scatter the client models). Extensive experiments show that FedSkip achieves significant improvements on FedAvg under the non-IID settings and outperforms a range of state-of-the-art methods with high communication efficiency and aggregation efficiency. In the future, it is worthwhile to explore more comprehensive combinations between federated learning methods and federated skip aggregation.

\section{ACKNOWLEDGMENT}
This work is supported by the National Key R\&D Program of China (No. 2019YFB1804304), 111 plan (No. BP0719010),  and STCSM (No. 18DZ2270700, No. 21DZ1100100), and State Key Laboratory of UHD Video and Audio Production and Presentation.

\bibliographystyle{IEEEtran}
\bibliography{main.bib}

\section*{Appendix}
The supplementary material provides the proof of the Theorem~\ref{thm:fedskip}. 

\subsection{Additional notations}\label{sec:addition}
Here we rewrite the local training by total local updates $\tau$,
$$
\v^{\tau}_{k}=\w^{\tau-1}_k - \eta_\tau \nabla F_k(b^{\tau-1}_k;\w^{\tau-1}_k),
$$
and
\begin{equation}
    \label{eq:init2}
    \w^{t}_{k}\leftarrow\left\{
    \begin{array}{ll}
\w^{0}, & t=0 \\
{\hat{\w}^{t-1}_{k}} ,  & t~\in~I_{skip}\\
\sum_{k=1}^{K}p_k^t\w_{k}^{t-1}, & \text{otherwise} \\
\end{array} \right.
\end{equation}
We define $\aw^\tau = \myave \aw^\tau_k$ and $\av^\tau = \myave \av^\tau_k$. They satisfy
\begin{small}
\begin{align*}\small
&\aw^{\tau+1} = \aw^\tau - \eta_\tau \g_\tau \\
&\av^{\tau+1} = \aw^{\tau+1}
\end{align*}
\end{small}
, where $\g_\tau = \myave \nabla F_k(\w^\tau_k, b^\tau_k)$, $\EB \g_\tau = \overline{\g}_\tau$ and $\overline{\g}_\tau = \myave \nabla F_k(\aw^\tau_k)$. 

\subsection{Proof of Theorem 1}\label{sec:proof}
Before proof our theorem, we first give some key lemmas that used as intermediate results. Lemma~\ref{lem:conv_main},~\ref{lem:conv_variance} have already been proved~\cite{convergence} and the Lemma~\ref{lem:math} is a math tool~\cite{stich}. So we only give the proof of the Lemma~\ref{lem:conv_diversity} in the APPENDIX C.
\begin{lemma}[Results of one step SGD]\label{lem:conv_main} 
Assume Assumption~\ref{asm:smooth_and_strong_cvx} holds. If $\eta_t\leq \frac{1}{4L}$, we have
\begin{small}
\begin{align*}
\EB \left\|\overline{\w}^{\tau+1}-\w^{\star}\right\|^{2} 
\leq& (1-\eta_{\tau} \mu) \EB \left\|\overline{\w}^{\tau}-\w^{\star}\right\|^{2}+ 6L \eta_{\tau}^2 \Gamma \\
&+ \eta_{\tau}^{2} \EB \left\|\mathbf{g}_{\tau}-\overline{\mathbf{g}}_{\tau}\right\|^{2} + 2 \EB \myave \left\|\overline{\w}^{\tau}-\w_{k}^{\tau}\right\|^{2}
\end{align*}
\end{small}
where $\Gamma = F^* - \myave F_k^{\star} \ge 0$.
\end{lemma}

\begin{lemma}[Bounding the variance]\label{lem:conv_variance}
Assume Assumption~\ref{asm:sgd_var_and_norm} holds. It follows that
\begin{small}
\[\EB \left\|\mathbf{g}_{\tau}-\overline{\mathbf{g}}_{\tau}\right\|^{2}  \leq \sum_{k=1}^N p_k^2 \sigma_
k^2.\]
\end{small}
\end{lemma}

\begin{lemma}[Bounding the divergence of $\{\w_k^{\tau}\}$]\label{lem:conv_diversity}
Assume Assumption~\ref{asm:sgd_var_and_norm}, that $\eta_t$ is non-increasing and $\eta_{\tau} \le 2 \eta_{\tau+\Delta E}$ for all $\tau\geq 0$. It follows that
\begin{small}
\[\EB \left[ \myave \left\|\overline{\w}^{\tau}-\w_{k}^{\tau}\right\|^{2} \right]
\: \leq \: 4\Delta\eta_\tau^2 E^2 G^2.\]
\end{small}
\end{lemma}

\begin{lemma}[Math tool from Stich\cite{stich,lemma4}]\label{lem:math}
Assume there are two non-negative sequences $\{r_\tau\}$, $\{s_\tau\}$ that satisfy the relation
\begin{small}
\[r_{\tau+1}\leq(1-\alpha\gamma_\tau)r_\tau-b\gamma_\tau s_\tau+c\gamma_\tau^2\]
\end{small}
for all $\tau\geq0$ and for parameters $b>0, a>0, c>0$ and non-negative step sizes $\{\gamma_\tau\}$ with $\gamma_\tau\leq\frac{1}{d}$ for a parameter $d\geq a, d>0$. Then, there exists weights $\omega_\tau\geq0, W_T:=\sum_{\tau=0}^{T}\omega_\tau$, such that:
\begin{small}
\[\frac{b}{W_T}\sum_{\tau=0}^{T}s_\tau\omega_\tau+a r_{T+1}\leq32d r_0\exp{\left[-\frac{a T}{2d}\right]}+\frac{36c}{aT}\]
\end{small}
\end{lemma}


From Lemma~\ref{lem:conv_main}, Lemma~\ref{lem:conv_variance}, Lemma~\ref{lem:conv_diversity} and the fact that $\aw^{\tau+1}= \av^{\tau+1}$, it follows
\begin{small}
\[  
\EB \left\|\overline{\w}^{\tau+1}-\w^{\star}\right\|^{2} \le (1-\eta_\tau \mu) \EB \left\|\overline{\w}^{\tau}-\w^{\star}\right\|^{2} + \eta_\tau^2 C\]
\end{small}
where
\begin{small}
\[ C = \sum_{k=1}^N p_k^2 \sigma_k^2 +  6L \Gamma + 8 \Delta E^2G^2. \]
\end{small}
Let $r_\tau=\EB \left\|\aw^{\tau}-\w^{\star}\right\|^{2}$. We have
\[r_{\tau+1}\le (1-\eta_\tau\mu )r_\tau + \eta_\tau^2 C.\]
By setting $s_\tau=0, a=\mu, c=C, \gamma_\tau^2=\eta_\tau^2, d=1$, we can obtain from Lemma~\ref{lem:math}:
\begin{small}
\[\EB \left\|\overline{\w}^{T+1}-\w^{\star}\right\|^{2} \le \frac{32}{\mu}(\w^{0}-\w^{\star})\exp{\left[-\frac{\mu T}{2}\right]}+\frac{36C}{\mu^2 T}.\]
\end{small}
Here we complete the proof.

\subsection{Proof of lemma 3}\label{sec:lemma3}
    For the convenience of the proof, we define $\tau_0=\tau\bmod{(E*\Delta)}$ is the local updates that nearest to the last aggregation and $\tau_1,\tau_2,\dots,\tau_s,\dots,\tau_\Delta$ are the local updates that satisfy $\tau_s=\tau_0+sE$, $\tau_0+sE\leq\tau$ and $\tau_0+(s+1)E\geq\tau$, where $\Delta$ is the number of skip between two aggregations. Therefore, for any $\tau \ge 0$, there exists a $\tau_0 \leq \tau$, such that $\tau-\tau_0 \leq \Delta E$ and $\w^{\tau_0}_k=\aw^{\tau_0}$ for all $k=1,2,\cdots,N$. Also, we use the fact that $\eta_\tau$ is non-increasing and assume $\eta_{\tau_0} \leq 2 \eta_\tau$ for all $\tau-\tau_0 \leq \Delta E-1$, then
\begin{small}
\begin{align*}
        &\EB  \myave \left\|\overline{\w}^{\tau}-\w_{k}^{\tau}\right\|^{2} \\&= \EB  \myave \left\|  (\w_{k}^{\tau} - \aw^{\tau_0})  - (\overline{\w}^{\tau} - \aw^{\tau_0})   \right\|^{2} \\
        &\leq \EB \myave \left\|  \w^{\tau}_{k} - \aw^{\tau_0}  \right\|^{2} \\
        & = \EB \myave \left\|  (\w^{\tau}_{k}-\w^{\tau_s}_{k})+(\w^{\tau_s}_{k}-\w^{\tau_{s-1}}_{k})+\dots+( \w^{\tau_1}- \aw^{\tau_0})  \right\|^{2} \\
        & \leq \EB \myave \left(\left\|  (\w^{\tau}_{k}-\w^{\tau_s}_{k})\right\|^{2}+\dots + \left\| (\w^{\tau_1}_{k}-\aw^{\tau_0}_{k})\right\|^{2} \right) \\
        & \leq \myave G^2\left(  (\tau-\tau_s)^{2}\eta_{\tau_s}^2+  (\tau_s-\tau_{s-1})^{2}\eta_{\tau_{s-1}}^2+\dots+   (\tau_1-\tau_0)^{2}\eta_{\tau_0}^2\right)\\
        & \leq 4 \Delta\eta_\tau^2 E^2 G^2.
\end{align*}
\end{small}
Here in the first inequality, we use $\EB \| X - \EB X \|^2 \le \EB \|X\|^2$ where $X = \w^\tau_{k} - \overline{\w}^{\tau_0}$ with probability $p_k$.
        In the second inequality, we use $\left\|a_1+a_2+\dots+a_n\right\|^2 \leq a_1^2+a_2^2+\dots+a_n^2$.
        In the third inequality, we use $\EB \|\nabla F_k(\w^\tau_{k},b^\tau
_k)  \|^2 \le G^2$ for $k=1,2,\cdots N$ and $\tau \ge 1$ and Jensen inequality:
\begin{small}
\[
\left\|  \w^{i}_{k} - \aw^{j}  \right\|^{2} = 
\left\|  \sum_{\tau=i}^{j} \eta_{\tau}  \nabla F_k(\w^\tau_{k},b^\tau_k)    \right\|^{2}
\le (i-j) \sum_{\tau=i}^{j} \eta_{\tau}^2 \left\|  \nabla F_k(\w^\tau_{k},b^\tau_k)   \right\|^2,
\]\end{small}
In the second last inequality, we use $\eta_{\tau_0} \le 2 \eta_{\tau}$ for all $\tau-\tau_0 \leq \Delta E-1$.
\end{document}